%% file: neurips_2025.tex
\documentclass{article}

\usepackage[preprint]{neurips_2025}

\usepackage[utf8]{inputenc} 
\usepackage[T1]{fontenc}    
\usepackage{hyperref}       
\usepackage{url}            
\usepackage{booktabs}       
\usepackage{amsfonts}       
\usepackage{nicefrac}       
\usepackage{microtype}      

\usepackage{amsmath}
\usepackage{graphicx}
\usepackage{longtable}  
\usepackage{multirow} 
\usepackage[table]{xcolor}
\usepackage{arydshln}
\usepackage{svg}
\usepackage{subcaption} 
\usepackage{caption} 
\usepackage{makecell} 
\usepackage{pifont}
\usepackage{cleveref}
\usepackage{url} 
\title{\textbf{BASE-Q: Bias and Asymmetric Scaling Enhanced Rotational Quantization for Large Language Models}
}

\author{%
  Liulu He \\
  Nanjing University\\
  \texttt{602022230015@smail.nju.edu.cn} \\
  \And
  Shenli Zheng \\
  Nanjing University \\
  \texttt{shenlizhen@smail.nju.edu.cn} \\
  \And
  Kaiwei Sun \\
  Nanjing University \\
  \texttt{522024230057@smail.nju.edu.cn} \\
  \And
  Yijiang Liu \\
  Nanjing University \\
  \texttt{yijiangliu@smail.nju.edu.cn} \\
  \And
  Yufei Zhao \\
  Nanjing University \\
  \texttt{yufeizhao@smail.nju.edu.cn} \\
  \And
  Chongkang Tan\\
  Alibaba Group \\
  \texttt{chongkangtan@antgroup.edu.cn} \\
  \And
  Huanrui Yang \\
  University of Arizona \\
  \texttt{huanruiyang@arizona.edu} \\
  \And
  Yuan Du \\
  Nanjing University \\
  \texttt{yuandu@nju.edu.cn} \\
  \And
  Li Du \\
  Nanjing University \\
  \texttt{ldu@nju.edu.cn} \\
}

\begin{document}

\maketitle

\begin{abstract}

  Rotations have become essential to state-of-the-art quantization pipelines for large language models (LLMs) by effectively smoothing outliers in weights and activations. 
  However, further optimizing the rotation parameters offers only limited performance gains and introduces significant training overhead: due to rotation parameter sharing, full-model must be loaded simultaneously to enable backpropagation, resulting in substantial memory consumption and limited practical utility.
  In this work, we identify two fundamental limitations of current rotational quantization methods: (i) rotation fails to align channel means, resulting in wider quantization bounds and increased rounding errors; and (ii) rotation makes the activation distribution more Gaussian-like, increasing energy loss caused by clipping errors. 
  To address these issues, we introduce \textbf{BASE-Q}, a simple yet powerful approach that combines bias correction and asymmetric scaling to effectively reduce rounding and clipping errors. Furthermore, BASE-Q enables blockwise optimization, eliminating the need for memory-intensive full-model backpropagation.
  Extensive experiments on various LLMs and benchmarks demonstrate the effectiveness of BASE-Q, narrowing the accuracy gap to full-precision models by 50.5\%, 42.9\%, and 29.2\% compared to QuaRot, SpinQuant, and OSTQuant, respectively. The code can be found at \url{https://github.com/Heliulu/BASE-Q}.

\end{abstract}

\section{Introduction}
Large language models (LLMs)\cite{touvron2023llama,Bai2023QwenTR,Dubey2024TheL3,Touvron2023Llama2O,DeepSeekAI2024DeepSeekV3TR} drive advances across diverse natural language tasks, but their ever-increasing scales present significant challenges for efficient deployment, particularly regarding inference latency and memory consumption. Low-precision quantization\cite{adaround,Li2021BRECQPT,Liu2022NoisyQuantNB,Liu2025FBQuantFQ} (e.g., 4 bits or lower) is a crucial technique for mitigating these bottlenecks. However, aggressively reducing numerical precision to such low bit-widths results in substantial accuracy degradation. A major challenge underlying this degradation is the presence of significant outliers in activations and weights\cite{Wei2023OutlierSA, Dettmers2022LLMint88M}, which necessitates wider quantization ranges and consequently amplifies quantization error.


To address the challenge of outliers in LLM quantization, linear equivalent transformations, such as scaling and rotation, have emerged.
For instance, SmoothQuant\cite{Xiao2022SmoothQuantAA} introduces channel-wise scaling to reduce activation variance, substantially smoothing distributions and delivering robust improvements in INT8 post-training quantization (PTQ) across various LLMs. Building on this direction, QuaRot\cite{Ashkboos2024QuaRotO4} introduces Hadamard rotations as another form of equivalent transformation to redistribute outliers and reduce channel-wise disparities, leading to robust INT4 PTQ. Since RMSNorm in LLMs normalizes activation energy for every residual block, and this energy is preserved under rotation, the same rotation matrix can be shared across all blocks and fused into adjacent linear layers, incurring no additional computation. 

Building upon the effectiveness of fixed rotations like Hadamard, subsequent research (e.g., SpinQuant\cite{Liu2024SpinQuantLQ}, OSTQuant\cite{Hu2025OstQuantRL}) has investigated optimizing more complex, learnable rotations for enhanced performance. While these approaches may offer marginal gains, they implicitly highlight that fixed rotations do not resolve all quantization error sources. Crucially, optimizing these learnable rotations requires computationally and memory-intensive full-model optimization procedures, as the rotation matrix is shared by all blocks, making them significantly less practical than fixed-rotation approaches. Therefore, to advance rotation-based quantization without impractical overhead, it is essential to (i) \textbf{identify the residual quantization errors} after fixed rotation and (ii) \textbf{propose efficient techniques} to mitigate these errors without relying on full-model optimization.


Motivated by these challenges, we conduct a detailed analysis of post-rotation quantization errors to identify the key bottlenecks in current rotation-based methods.
Firstly, our analysis reveals that even after applying rotation, a dominant source of rounding error stems from the variance of channel-wise means. While further optimizing rotation parameters can partially mitigate this issue, the objectives of outlier suppression and channel mean alignment are inherently conflicting—especially when a single rotation is shared across all blocks. This conflict severely constrains the rotation's ability to simultaneously achieve both goals, forcing suboptimal trade-offs.
To address the problem, we fix the rotation matrix shared across all blocks instead of training it, and introduce a channel bias correction mechanism that effectively aligns channel means with negligible inference overhead and lightweight block-wise optimization.
Secondly, we uncover the detrimental effect caused by dynamic activation clipping, which is commonly adopted to suppress outliers.
Rotation transforms distributions into more compact, near-Gaussian shapes.
Consequently, quantization clipping impacts a larger fraction of activation values, significantly reducing their total energy and fundamentally breaking the computational equivalence established by rotation. To address this, we introduce an asymmetric scaling strategy: by appropriately increasing the pre-quantization scaling factor, we preserve the activation energy after clipping.

Our main contributions are:
\begin{itemize}
    \item \textbf{Theoretical Insight}. We provide a systematic analysis that reveals key components of quantization error after rotation, uncovering why existing approaches plateau and where further optimization is possible.
    \item \textbf{Efficient Blockwise Correction}. We propose BASE-Q, a novel and practical quantization framework that uses blockwise channel bias correction and asymmetric scaling under fixed rotations, achieving significant performance enhancement without the expansive full-model optimization.
    \item \textbf{Extensive Empirical Validation}. Extensive experiments on various LLMs and benchmarks demonstrate the effectiveness of our BASE-Q, narrowing the accuracy gap to full-precision models by 50.5\%, 42.9\%, and 29.2\%  compared to QuaRot\cite{Ashkboos2024QuaRotO4}, SpinQuant\cite{Liu2024SpinQuantLQ}, and OSTQuant\cite{Hu2025OstQuantRL}, respectively.
\end{itemize}

\section{Related work}
\textbf{Equivalent Transformations in LLM Quantization.} Post-training quantization (PTQ) has attracted considerable attention for the efficient deployment of LLMs, but remains challenging due to frequent outliers in both activations and weights. 
\textbf{AWQ}\cite{lin2024awq} introduced channel-wise scaling for weight-only PTQ, while \textbf{SmoothQuant}\cite{Xiao2022SmoothQuantAA} tailored rescaling for activations and weights to suppress outlier effects and enable robust INT8 quantization. \textbf{OmniQuant}\cite{Shao2023OmniQuantOC} extended this concept with learnable scaling coefficients for each submodule, allowing finer-grained adaptation across network components. \textbf{AffineQuant}\cite{Ma2024AffineQuantAT} further generalized these ideas by employing learnable affine transformations to jointly align the mean and variance before quantization. Beyond scaling-based methods, \textbf{QuIP}\cite{chee2023quip,tseng2024quip} first applied rotation transformations for weight-only PTQ. \textbf{QuaRot}\cite{Ashkboos2024QuaRotO4} proposed applying Hadamard rotations to both activations and weights, making distributions more Gaussian and further suppressing outliers, thus simplifying the quantization process. 
\textbf{DuQuant}\cite{lin2024duquant} employed rotation and permutation to more effectively eliminate outliers. 
\textbf{SpinQuant}\cite{Liu2024SpinQuantLQ} took this further by learning optimal rotation matrices from calibration data, achieving lower quantization errors at the cost of greater computational and memory requirements. \textbf{OSTQuant}\cite{Hu2025OstQuantRL} unified learnable rotations and scaling within a single framework, providing additional flexibility and consistently outperforming previous methods across various LLM benchmarks. 
\textbf{FlatQuant}\cite{sun2024flatquant} employed layer-wise learned online matrix transforms to improve quantized linears, at the cost of increased inference overhead and parameter count. 
Our theoretical insights enable BASE-Q to outperform previous rotational quantization baselines with only fixed rotation and no global optimizations.
\section{Error analysis of existing methods}
In this section, we theoretically analyze the errors derived by rotation-based quantization, specifically focusing on the Hadamard transformation. We begin by investigating the error reduction mechanism inherent in this transformation, deriving quantitative expressions for both rounding and clipping errors in \Cref{sec:qe-hadamard}. Subsequently, in \Cref{sec:round-error,sec:clip-error}, we provide an in-depth analysis of these error sources, uncovering potential components that are amenable to further optimization and providing key insights for methodological improvements. Finally, drawing inspiration from OSTQuant\cite{Hu2025OstQuantRL}, which demonstrates complementary effects of scaling and rotation, we interpret and substantiate this synergy through our error decomposition framework in \Cref{sec:scaling}.

\subsection{Quantization error analysis for Hadamard rotations}
\label{sec:qe-hadamard}
The key challenge in the LLM quantization is the presence of extreme outliers in activation channels, which significantly increases the dynamic range and severely raises quantization difficulty.
To clarify how Hadamard rotations mitigate this issue, consider a token activation $\boldsymbol{\mathit{x}}\in\mathbb{R}^n$ with outlier values structured as follows: 
\begin{equation}
    \boldsymbol{\mathit{x}}=\boldsymbol{\mathit{g}}+\sum_{i\in\mathcal{O}}a_i\delta \boldsymbol{\mathit{e}}_i ,
\end{equation}
where $\boldsymbol{\mathit{g}}\sim \mathcal{N}(\mu, \delta^2\mathbf{I})$ represents the “main mass” as a Gaussian component, and each outlier channel $i \in \mathcal{O}$ is modeled as a scaled one-hot vector of amplitude $a_i\delta$ with $a_i\gg1$ and $|\mathcal{O}|\ll n$.
When an orthogonal Hadamard rotation $\boldsymbol{\mathit{H}}$ is applied (where each entry is $\pm n^{-\frac{1}{2}}$), the transformed activation becomes: 
\begin{equation}
    \boldsymbol{\mathit{Hx}}=\boldsymbol{\mathit{Hg}}+\sum_{i\in\mathcal{O}}a_i\delta \boldsymbol{\mathit{He}}_i .
\end{equation}
Since Gaussian distributions are invariant under orthogonal transformations, the main mass $\boldsymbol{\mathit{Hg}}$ remains Gaussian. Critically, each outlier term $a_i\delta\boldsymbol{\mathit{e}}_i$ is now mapped to the $i$-th column of $\boldsymbol{\mathit{H}}$, a dense vector whose large amplitude $a_i\delta$ is evenly spread across all channels, with each channel receiving only 
$a_i\delta/\sqrt{n}$. For large $n$, the per-channel impact of any outlier is significantly diminished; effectively, outliers are “absorbed” into the Gaussian bulk, yielding a distribution that is far more conducive to quantization.

This mechanism extends beyond simple Gaussian activations to more general cases where activations exhibit correlated structures or multiple modes (e.g., Gaussian mixtures or channel-wise mean/variance discrepancies).
In such cases, applying a principal component transformation to diagonalize the covariance matrix, followed by a Hadamard rotation, further balances the marginal variances and minimizes the prominence of outliers:
\begin{equation}
    \boldsymbol{\mathit{x}}'=\boldsymbol{\mathit{HU}}^{T}(\boldsymbol{\mathit{x}}-\mathbb{E}[\boldsymbol{\mathit{x}}]) ,
\end{equation}
where $\boldsymbol{\mathit{U}}$ is the principal component basis.
To quantify the effect on quantization, consider uniform quantization with per-channel clipping:
\begin{equation}
    \boldsymbol{\mathit{x}}_q=F_{clip}(F_{round}(\frac{\boldsymbol{\mathit{x}}-z}{s}),0,2^b-1)*s+z ,
\end{equation}
where $z$ and $s$ denote the quantization lower bound and step size, respectively, and $b$ is the target bit-width. The overall quantization error consists of two components:
\begin{itemize}
\setlength{\itemindent}{-0pt}
    \item \textbf{Rounding error}, with expected $\ell _2$ energy per channel: $\mathbb{E}[\|\varepsilon_{\text{round}}\|_2^2] = \frac{s^2}{12}$.
    \item \textbf{Clipping error}, defined by the mass outside quantization bounds: $\mathbb{E}[\|\varepsilon_{\text{clip}}\|_2^2] = \int^{z}_{-\infty}x^2P(x)dx  +\int^{+\infty}_{z+\Delta} x^2P(x)dx$, where $\Delta = s(2^b-1)$ and $P(x)$ is the empirical channel distribution.
\end{itemize}
By dispersing outlier energy and compressing the activation dynamic range, Hadamard transformations significantly reduce rounding errors by enabling smaller quantization steps $s$.

\subsection{Rounding errors from misaligned channel means}
\label{sec:round-error}
\begin{figure}[htbp]
  \centering
  \begin{subfigure}[b]{0.26\textwidth}
    \centering
    \includegraphics[width=\textwidth]{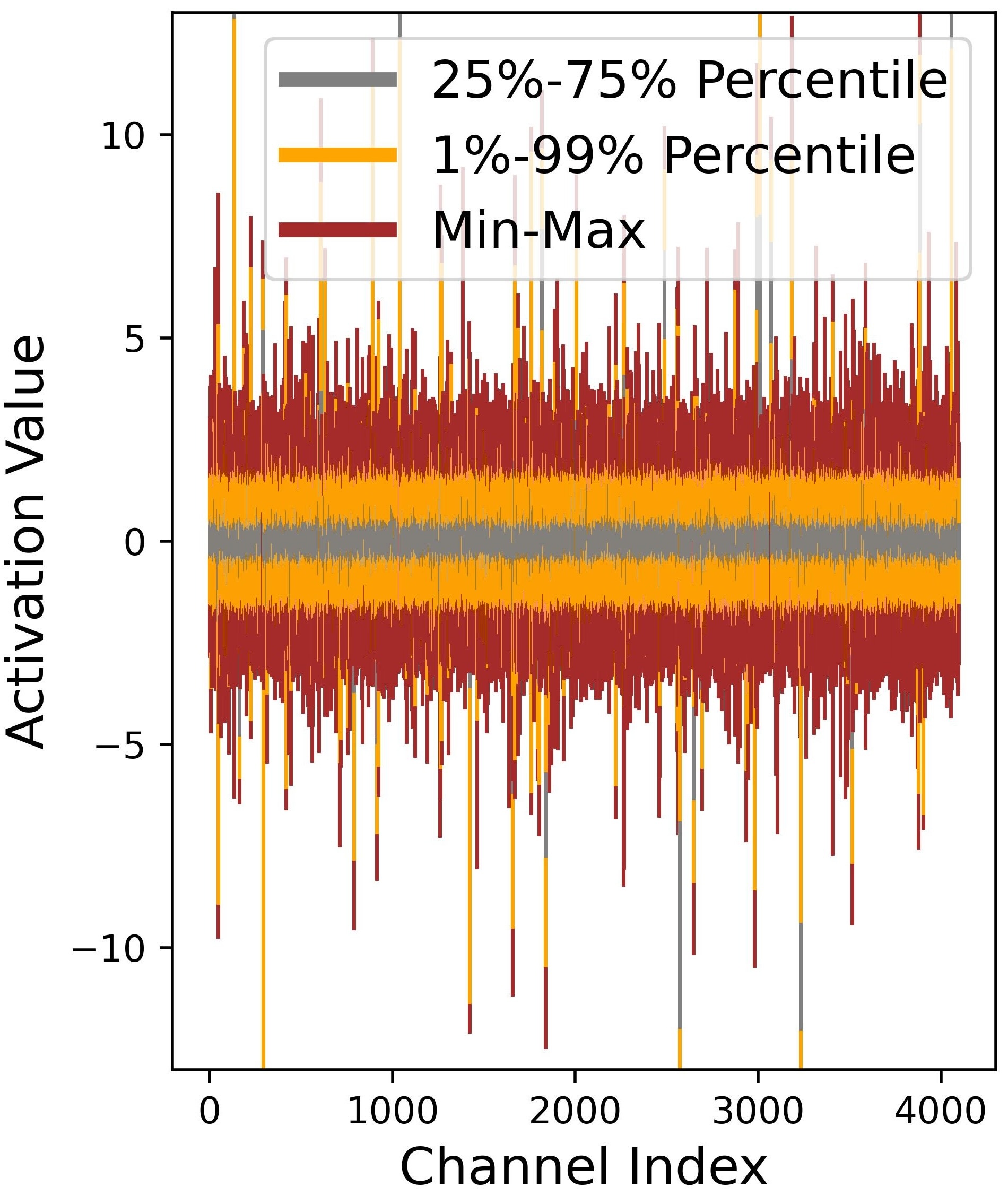}
    \caption{Raw Activation}
    \label{fig:sub-a}
  \end{subfigure}
  \hfill
  \begin{subfigure}[b]{0.24\textwidth}
    \centering
    \includegraphics[width=\textwidth]{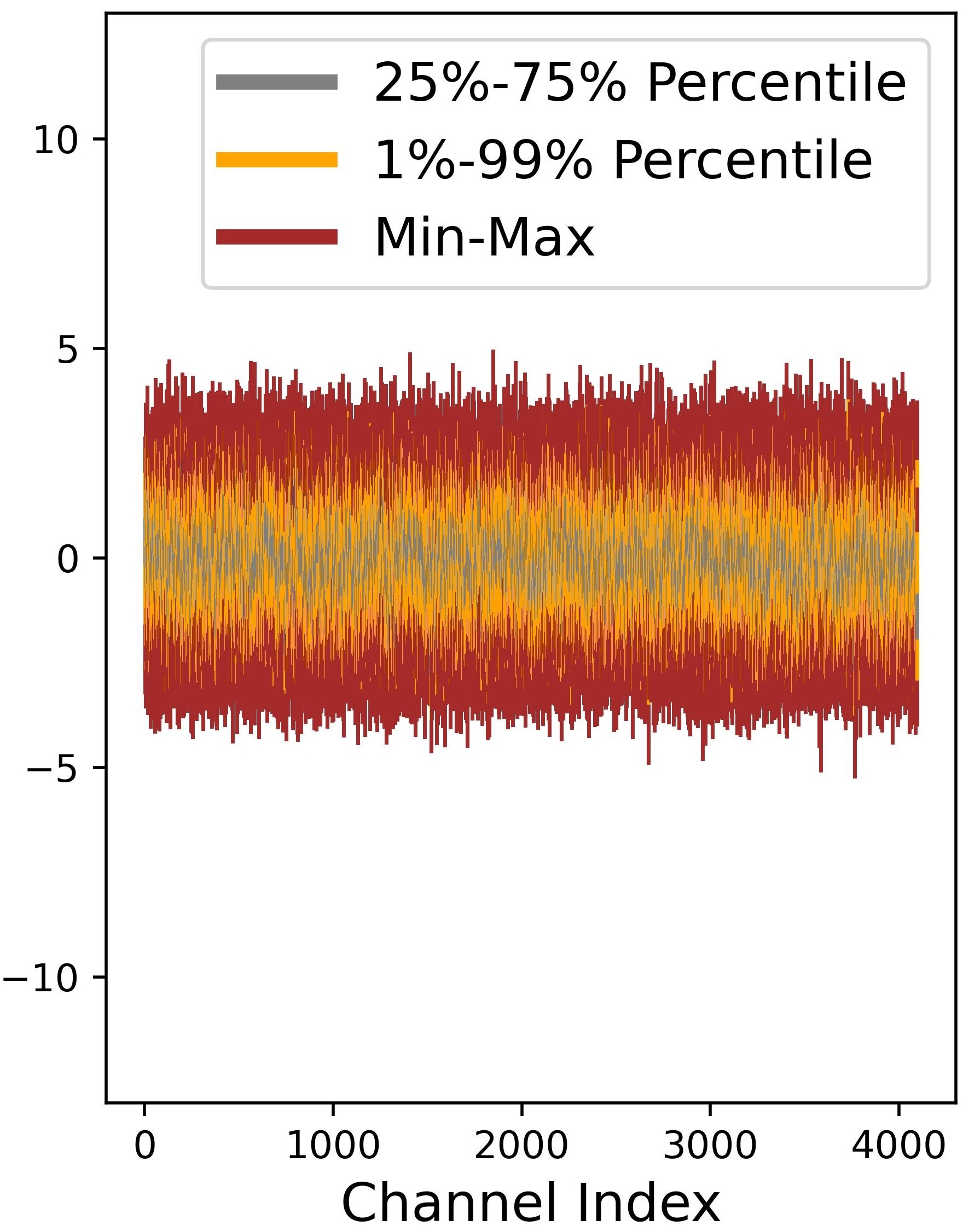}
    \caption{Hadamard $R_{res}$}
    \label{fig:sub-b}
  \end{subfigure}
  \hfill
  \begin{subfigure}[b]{0.24\textwidth}
    \centering
    \includegraphics[width=\textwidth]{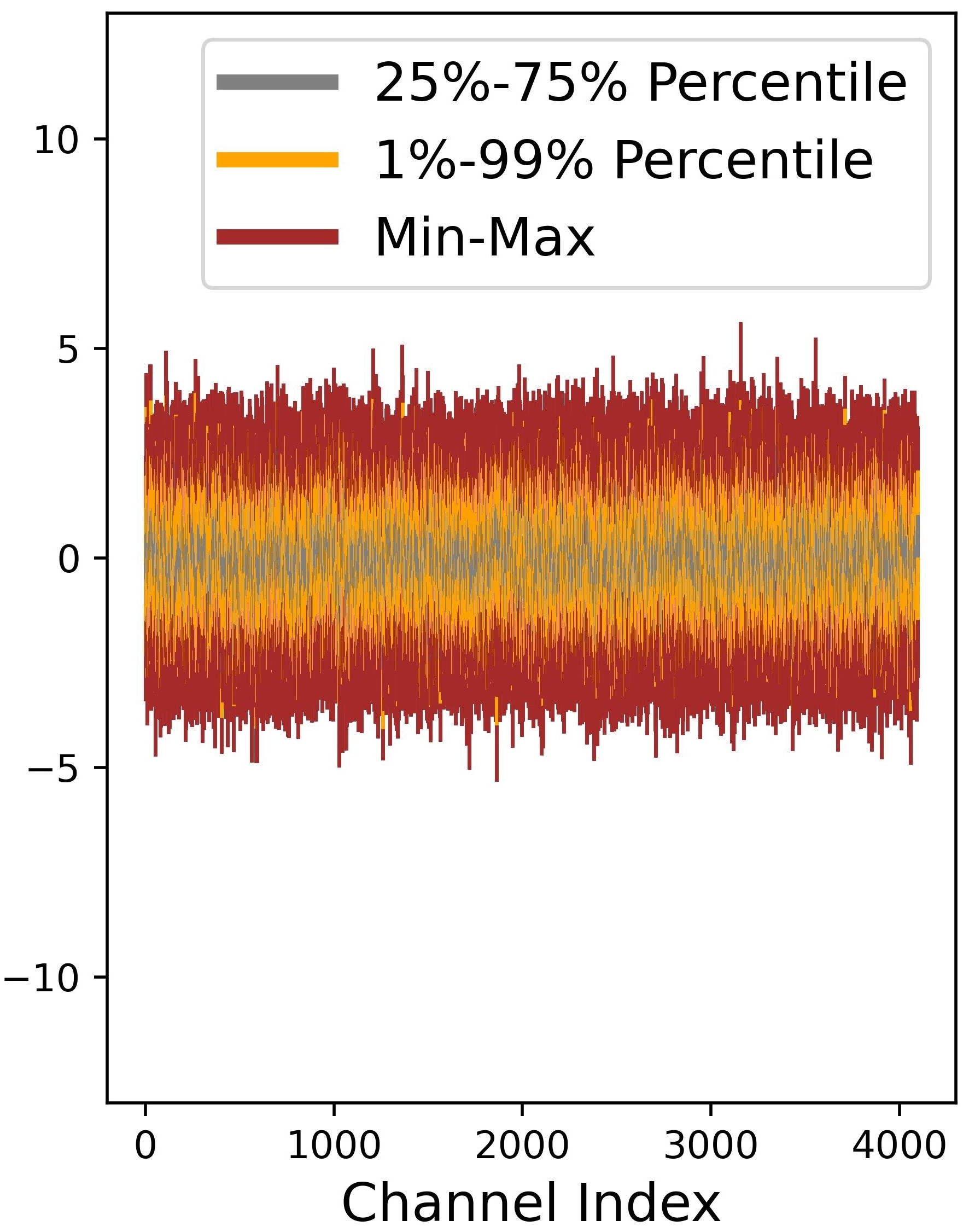}
    \caption{Learned $R_{res}$}
    \label{fig:sub-c}
  \end{subfigure}
  \hfill
  \begin{subfigure}[b]{0.24\textwidth}
    \centering
    \includegraphics[width=\textwidth]{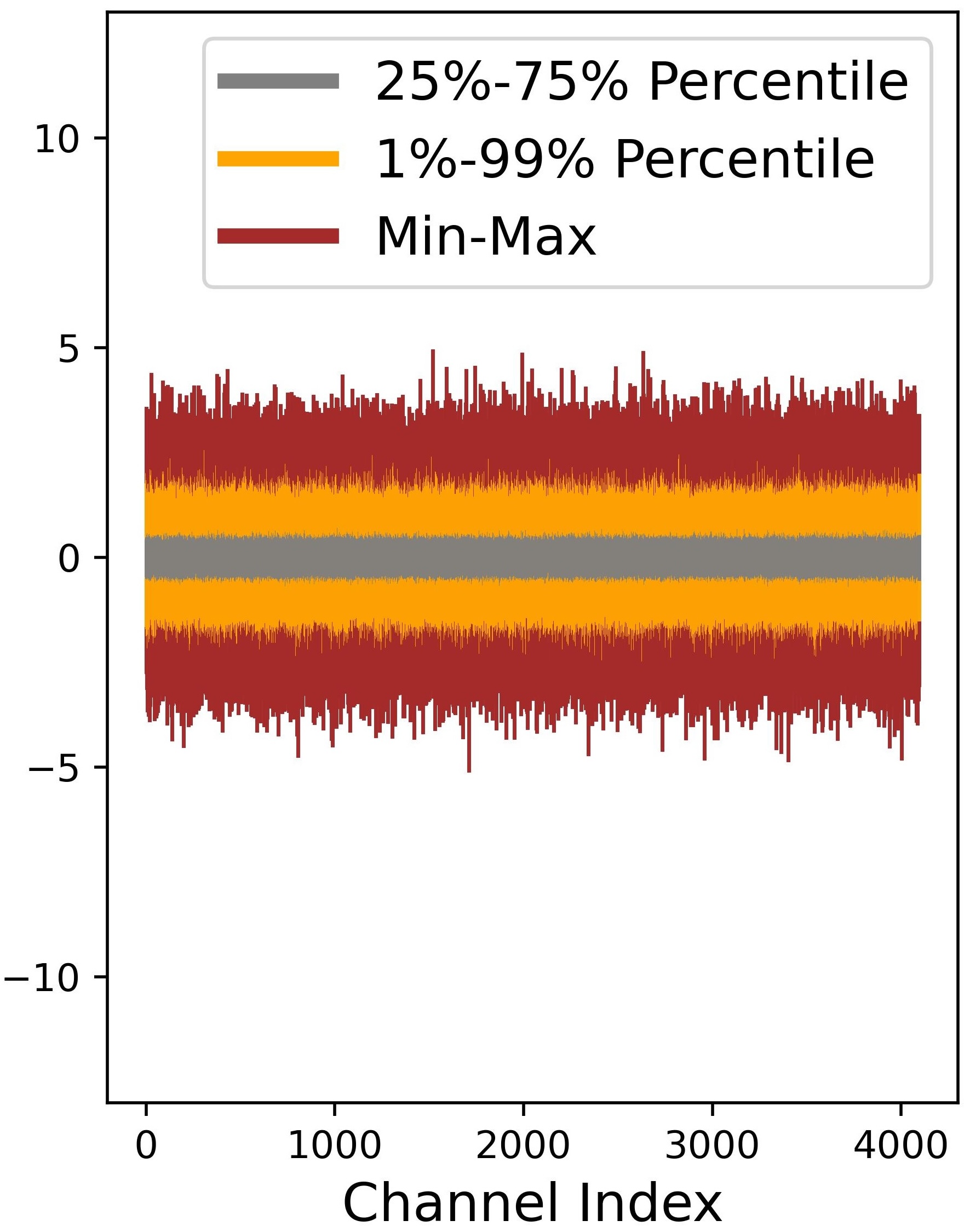}
    \caption{Hadamard + Bias Corr.}
    \label{fig:sub-c}
  \end{subfigure}
  \caption{(a) Raw activation distribution of the first MLP block in Llama3-8B. (b) Hadamard rotation suppresses outliers but leaves residual inter-channel misalignment. (c) Learned rotations also fail to address this misalignment. (d) Our bias correction eliminates most inter-channel mean variability, thereby reducing $Var(\mu_j)$, which is the main rounding error component as formalized in \Cref{eq::5,eq::6}. More visualizations are in \Cref{appendix:visual}.}
  \label{fig::main}
\end{figure}
As shown in the previous section, the energy of the rounding error is proportional to the quantization scale, $s = \frac{\Delta}{2^b-1}$. As rotations gaussianize the activation distribution, the quantization range can be seen as being proportional to the standard deviation of activations, $\sigma$. Therefore, the expected rounding error satisfies:
\begin{equation}
    \mathbb{E}[\|\varepsilon_{\text{round}}\|_2^2] \propto s^2 \propto \sigma^2 .
    \label{eq::5}
\end{equation}
The total variance $\sigma^2$ of the activations  can be decomposed into two components:
\begin{equation}
\sigma^2= \frac{1}{n}\sum^n_{j=1}\sigma ^2_j + Var(\mu_j) ,
\label{eq::6}
\end{equation}
where $\sigma ^2_j$ and $\mu_j$ are the variance and mean of the $j$-th channel, respectively. The first term, the average channel variance, is \textbf{invariant} under orthogonal transformations and dictated by the data distribution. 
In contrast, the second term, $Var(\mu_j)$ —the variance of channel means—can remain substantial in LLMs that use \textbf{RMSNorm}, since this normalization does not enforce alignment across channel means. As shown in \Cref{fig::main}, hadamard rotations are effective at attenuating outliers but leave the means of individual channels misaligned. 
Surprisingly, according to \Cref{eq::6}, we found that $Var(\mu_j)$ causes \textbf{up to 85\%} layer rounding error in Qwen2.5-3B (as shown in \Cref{fig:roundingerror}; additional details for more models are provided in \Cref{appendix:rounding error}).

Ideally, optimizing rotation would bring all channel means into alignment, achieving $Var(\mu_j)=0$. In practice, however, the LLM shares a single rotation $\boldsymbol{\mathit{R}}_{res}$ across all transformer blocks, despite block-wise variation in bias. The limited expressiveness of a global rotation makes it infeasible to simultaneously eliminate outliers and achieve mean alignment. 
As a result, globally learned rotations (e.g., SpinQuant) can only minimize $Var(\mu_j)$ in a least-loss sense but cannot completely remove this error from mean discrepancies. This motivates our explicit bias correction strategy, which precisely cancels the variance-of-means term post-rotation. Moreover, this approach supports blockwise optimization, avoiding the need for costly full-model optimization.

\subsection{Clipping eliminates non-negligible energy}
\label{sec:clip-error}

Orthogonal rotation, such as Hadamard transform, significantly alters the distribution of activations. While raw activations often exhibit a heavy-tailed distribution with sparse outliers, post-rotation activations approximate a Gaussian distribution. This transformation is critical: unlike heavy-tailed distributions where extreme values are rare, the Gaussian shape concentrates a larger proportion of activation energy towards the distribution tails. Consequently, applying a fixed clipping threshold removes a non-negligible amount of energy associated with values beyond the threshold.

\begin{minipage}{0.33\linewidth}
  \centering
  \includegraphics[width=\linewidth]{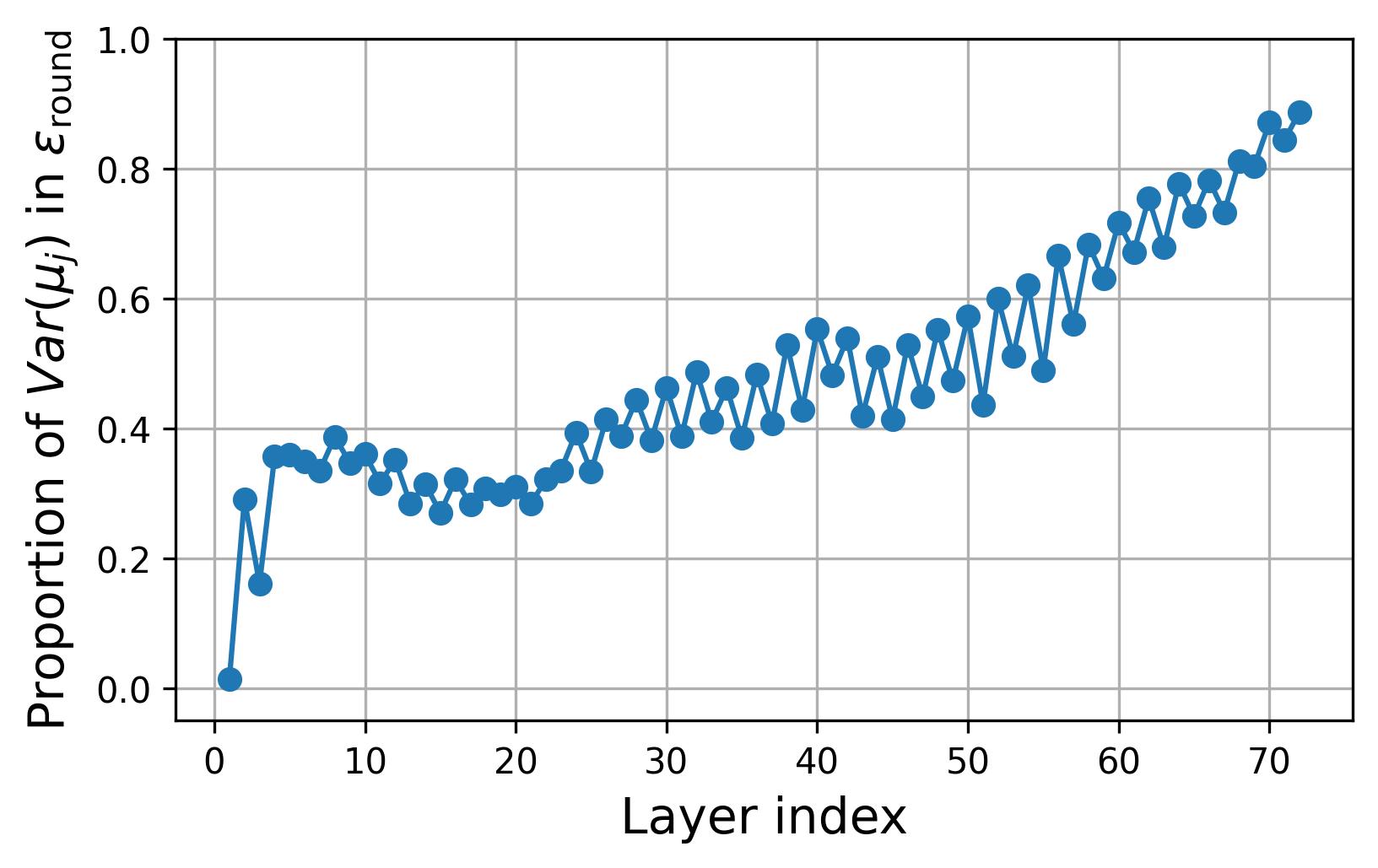}
  \captionof{figure}{Misalignment causes up to 85\% layer-wise rounding error in Qwen2.5-3B.}
  \label{fig:roundingerror}
\end{minipage}
\hfill
\begin{minipage}{0.65\linewidth}
  \centering
  \includegraphics[width=\linewidth]{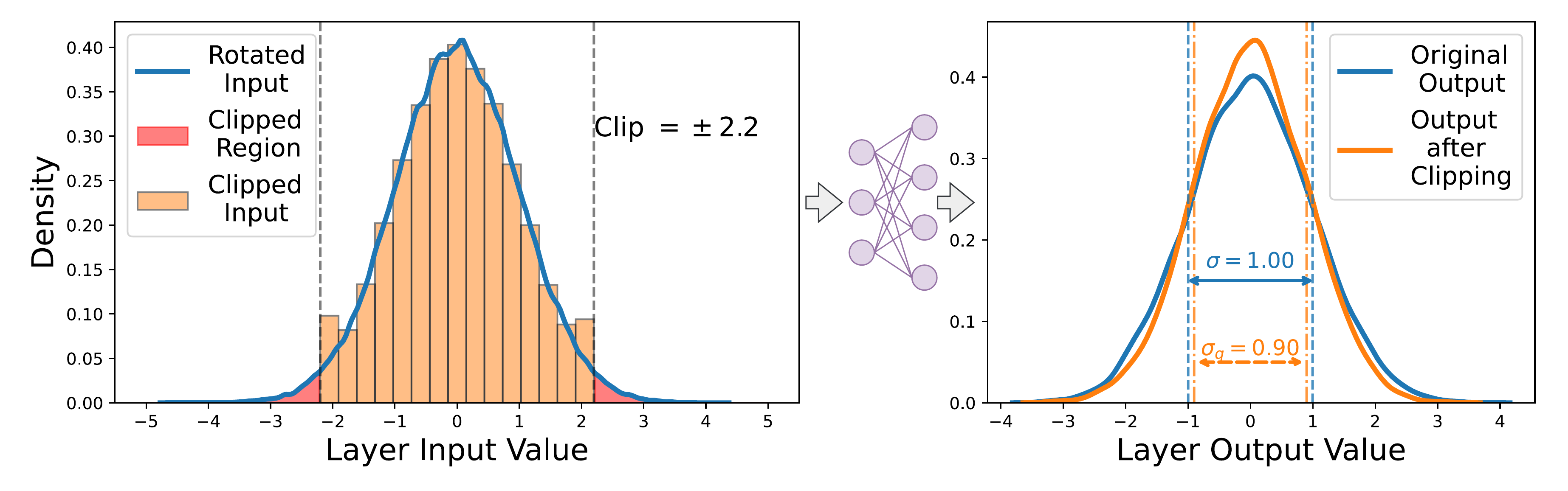}
  \captionof{figure}{(Left) MSE-optimal clipping on rotated activations results in a 18.4\% loss of layer's input energy, (Right) which induces significant discrepancies in the layer output.}
  \label{fig::symmetry}
\end{minipage}

Formally, consider activations $\boldsymbol{\mathit{x}}\sim \mathcal{N}(\mu,\sigma^2)$ quantized to $b=4$ bits (INT4), the minimum MSE-optimal clipping threshold is commonly found to be:
\begin{equation}
\theta^*=\underset{\theta}{\arg\min} \ \mathbb{E}[\|\boldsymbol{\mathit{x}}-\theta-F_{clip}(F_{round}(\frac{\boldsymbol{\mathit{x}}-\theta}{2\theta/(2^b-1)}),0,2^b-1)\|_2]\approx 2.2\sigma
\end{equation}
At this threshold, the proportion of the total activation energy contained within the clipped regions (i.e., the energy loss) can be quantified. For a zero-mean Gaussian, the expected $L_2$ energy of the clipped values is approximately 18.4\% of the total energy:
\begin{equation}
    \mathbb{E}[\| \varepsilon_{\text{clip}} \|_2] = \int^{-2.2\sigma}_{-\infty}x^2P(x)dx  +\int^{+\infty}_{+2.2\sigma} x^2P(x)dx \approx 18.4\% ,
\end{equation}
where $P(x)$ is the probability density function.
18.4\% of the energy is \textbf{non-negligible} in the context of deep neural networks. When this clipping is applied layer by layer throughout a deep network, the accumulated energy loss becomes substantial. The empirical consequence of this breakdown is illustrated in \Cref{fig::symmetry}, where the distribution of the layer output significantly mismatches the original.
Moreover, in practical LLM architectures, using globally or per-layer optimized clipping bounds cannot fully resolve this energy loss, as joint optimization across all layers is intractable. This limitation explains the observed diminishing returns of aggressive clipping in rotation-based quantization schemes, even with parameter tuning.

To address this issue, we introduce asymmetric scaling for each activation quantization step, which restores the energy loss and maintains accurate signal magnitude, thereby improving both theoretical fidelity and empirical performance.


\subsection{Role of scaling in rotational quantization}
\label{sec:scaling}
Combining scaling with orthogonal transformations (rotations) has been shown to be crucial for enhancing quantization performance.
Theoretically, the rounding noise introduced by quantization after rotation can be modeled as zero-mean uniform noise with variance determined by the quantization bin width, defined as:
\begin{equation}
    \epsilon \sim\mathcal{U}(-\frac{s}{2},\frac{s}{2}) ,
\end{equation}
where $\epsilon$ and $s$ denote the rounding noise and quantization bin width, respectively.
An orthogonal transform preserves the variance. Thus, the noise is equivalent to additive isotropic Gaussian noise in the original space, which becomes:
\begin{equation}
    \epsilon \sim\mathcal{N}(0,\frac{s^2}{12}) .
\end{equation}
Referring to the matrix multiplication of weights and activations in the specific layer, the layer output includes terms arising from error propagation:
\begin{equation}
    (w+\epsilon_w)\cdot(a+\epsilon_a) = w\cdot a + (w\cdot\epsilon_a + \epsilon_w \cdot a) + \epsilon_w \cdot \epsilon_a ,
\end{equation}
where $w$, $a$ are the original weights and activations, respectively, and $\epsilon_a$, $\epsilon_w$ are the corresponding quantization errors.

Introducing per-channel scaling factors $s$ (for $w$) and $1/s$ (for a) allows balancing the propagated error energy between these terms. The total variance of the noisy product, dominated by$w\cdot\epsilon_a$ and $\epsilon_w \cdot a$, is minimized when their variances are equal. By the arithmetic-geometric mean (AM-GM) inequality, the optimality is achieved if
\begin{equation}
s^2 = \frac{\mathbb{E}[|w|_2]}{\mathbb{E}[|a|_2]}.
\end{equation}
This choice of scaling enables optimal allocation of quantization noise energy, ensuring minimal loss in the presence of channel-wise variations and reinforcing the theoretical motivation for scaling in rotational quantization.

\section{BASE-Q}
As discussed in \Cref{sec:round-error,sec:clip-error}, while rotation-based quantization methods effectively smooth outliers, they exhibit three notable limitations:
(a) Channel-mean variance cannot be eliminated via rotation-only optimization, leading to a persistent rounding error term (as formalized in \Cref{eq::5,eq::6}).
(b) Implementing flexible and optimal activation clipping schemes within standard rotation-based PTQ is challenging, resulting in cumulative energy loss across layers.
(c) Optimizing full-network rotation parameters requires prohibitive GPU memory and computational resources, especially for large models.
To address these issues, we propose \textbf{BASE-Q} (\textbf{B}ias and \textbf{A}symmetric \textbf{S}caling \textbf{E}nhanced \textbf{Q}uantization), a lightweight yet powerful quantization framework that strengthens rotation-based methods through two key innovations: explicit \textbf{bias correction} to fully remove channel mean variance, and \textbf{asymmetric scaling} to improve clipping and quantizer fit at each block. The blockwise optimization capability  avoids the heavy cost and complexity of full-model rotation learning, thereby enabling efficiency and enhanced overall quantization performance.

\begin{figure} [h]
	\begin{center}
	\includegraphics[width=1\textwidth]{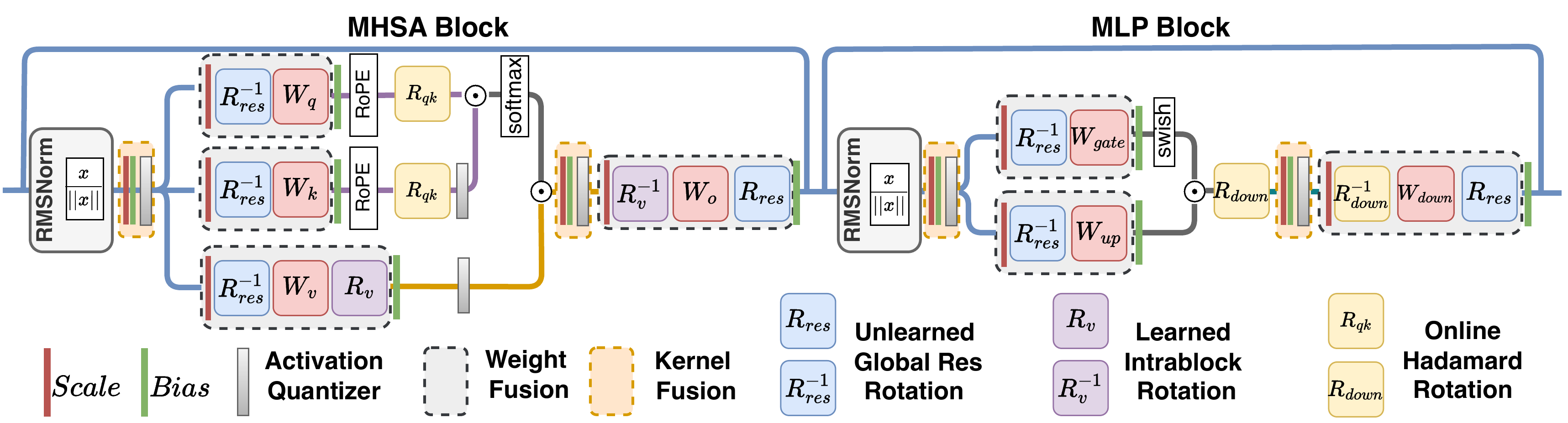}
	\caption{An overview of \textbf{BASE-Q}, highlighting three key design features: (a) it incorporates channel-wise bias correction to reduce rounding error in activation quantization. 
    (b) it applies asymmetric scaling to compensate for the loss of computational equivalence caused by the clipping; (c) it avoids learning $\boldsymbol{\mathit{R}}_{res}$, thereby eliminating full model optimization. This mechanism matches the parameter count and hardware fusion efficiency of typical scaling strategies.} 
	\label{fig::method}
        \end{center}
\end{figure} 

\paragraph{Fundamental Rotation Settings.} \Cref{fig::method} provides a schematic of our BASE-Q. The core optimized transformations within each transformer block include the global shared residual rotation ($\boldsymbol{\mathit{R}}_{res}$), the blockwise value rotation ($\boldsymbol{\mathit{R}}_v$), and the online rotations ($\boldsymbol{\mathit{R}}_{qk}$ and $\boldsymbol{\mathit{R}}_{down}$).
$\boldsymbol{\mathit{R}}_{res}$ is applied on the residual pathway, which can be fused into linear weights without extra inference overhead. Because the residual path connects all blocks, a single shared rotation is used throughout. As our method further removes channel mean variance and optimizes quantization for activations, it is preferable to set $\boldsymbol{\mathit{R}}_{res}$ to primarily optimize weight statistics. Following \Cref{sec:qe-hadamard}, we compute the PCA transform $\boldsymbol{\mathit{U}}$ across all related linear weights, and compose it with the Hadamard transformation:
\begin{equation}
   \boldsymbol{\mathit{R}}_{res}=\boldsymbol{\mathit{U}}^T\boldsymbol{\mathit{H}} ,
\end{equation}
where $\boldsymbol{\mathit{U}}$ is derived from the covariance of all associated linear weights.
$\boldsymbol{\mathit{R}}_v$ can be exactly fused layer-wise into linear parameters and supports independent, block-level optimization. We use PyTorch’s built-in Cayley orthogonal mapping for rotation learning, initialized as a Hadamard matrix.
$\boldsymbol{\mathit{R}}_{qk}$ and $\boldsymbol{\mathit{R}}_{down}$ require on-the-fly computation during inference. To reduce memory and computation overhead, we adopt Hadamard rotations, which have binary elements and can be efficiently implemented through fast algorithms. This setup matches the conventions of leading baselines such as QuaRot\cite{Ashkboos2024QuaRotO4}, SpinQuant\cite{Liu2024SpinQuantLQ}, and OSTQuant\cite{Hu2025OstQuantRL}.

\paragraph{Bias Correction.} To target the critical channel-mean variance term, we inject a learnable bias before activation quantization. After quantized inference and the corresponding linear projection, this bias is subtracted, yielding:
\begin{equation}
   \boldsymbol{\mathit{y}}= Q_w(\boldsymbol{\mathit{WR}})\;Q_a([\boldsymbol{\mathit{R}}^{-1}\boldsymbol{\mathit{x}}-\boldsymbol{\mathit{b}}^c])+\underset{fused\;bias}{\underbrace {\boldsymbol{\mathit{WRb}}^c+\boldsymbol{\mathit{b}}}}
\end{equation}
where $\boldsymbol{\mathit{WR}}$ denotes rotated weights and $b$ is the usual layer bias. The learnable bias correction terms—including $\boldsymbol{\mathit{b}}^c_{qkv}$, $\boldsymbol{\mathit{b}}^c_{o}$, $\boldsymbol{\mathit{b}}^c_{up}$ and  $\boldsymbol{\mathit{b}}^c_{down}$, which incur negligible parameter overhead (less than 0.1\% of original model size), and are highly efficient to optimize.

\paragraph{Asymmetric Scaling.} For the \textit{O} and \textit{Down} linear layers, we introduce both global symmetric $\boldsymbol{\mathit{s}}$ and per-quantizer asymmetric scaling $\boldsymbol{\mathit{s}}^a$. Asymmetric scaling $\boldsymbol{\mathit{s}}^a$ can adjusts the quantization range at inference time to better fit activation statistics and clipping needs:
 \begin{equation}
   Y =  Q_w(\boldsymbol{\mathit{WsR}})\; Q_a(\boldsymbol{\mathit{s}}^a\boldsymbol{\mathit{R}}^{-1}\boldsymbol{\mathit{s}}^{-1}\boldsymbol{\mathit{x}}-\boldsymbol{\mathit{b}}^c)+(\boldsymbol{\mathit{WRb}}^c+\boldsymbol{\mathit{b}})
\end{equation}

\paragraph{Blockwise Optimization.} Combining these techniques, our blockwise optimization strategy jointly optimizes $\boldsymbol{\mathit{R}}_v$, $\boldsymbol{\mathit{b}}^c_i$,$\boldsymbol{\mathit{s}}_j$, $\boldsymbol{\mathit{s}}^a_i$, and activation clipping threshold $\alpha_i$ per block via minimizing the MSE between the floating-point and quantized outputs:
\begin{equation}
    \underset{\boldsymbol{\mathit{R}}_v, \boldsymbol{\mathit{b}}^c_i,\boldsymbol{\mathit{s}}_j, \boldsymbol{\mathit{s}}^a_i, \alpha_k}{argmin}\  \mathcal{L}_{mse}(\boldsymbol{\mathit{y}}_{FP},\boldsymbol{\mathit{y}}_{Q};\boldsymbol{\mathit{R}}_v, \boldsymbol{\mathit{b}}^c_i,\boldsymbol{\mathit{s}}_j, \boldsymbol{\mathit{s}}^a_i, \alpha_i, \theta)
\end{equation}
where $\boldsymbol{y}_{FP}$ and $\boldsymbol{y}_{Q}$ denote the reference and quantized outputs, and $\theta$ is the set of frozen model and rotation parameters. This framework achieves state-of-the-art quantization accuracy with negligible memory and compute overhead under quantization process. 

\section{Experiment}
\label{sec:experiment}
\paragraph{Models and Tasks.} We evaluate our method on 12 open-source LLMs, spanning various sizes: Llama-2-7/13/70B\cite{touvron2023llama}, Llama-3-8/70B\cite{Dubey2024TheL3}, Llama-3.1-8/70B, Llama-3.2-1/3B, and Qwen2.5-3/14/32B\cite{yang2024qwen2}. We report perplexity on the Wikitext-2 dataset\cite{merity2016pointer} and zero-shot accuracy on nine tasks, including ARC-Easy and ARC-Challenge\cite{clark2018think}, BoolQ\cite{clark2019boolq}, HellaSwag\cite{zellers2019hellaswag}, LAMBADA\cite{radford2019language}, OpenBookQA\cite{mihaylov2018can}, PIQA\cite{bisk2020piqa}, SIQA\cite{sap2019socialiqa}, and WinoGrande\cite{sakaguchi2021winogrande}. Baseline methods include QuaRot\cite{Ashkboos2024QuaRotO4}, which uses fixed Hadamard rotations, SpinQuant\cite{Liu2024SpinQuantLQ}, which leverages learned rotations, and OSTQuant\cite{Hu2025OstQuantRL}, which employs both learned rotations and scaling.

\paragraph{Deployment Details.} Our quantization is implemented in PyTorch and evaluated using lm-eval\cite{eval-harness}. Activations are quantized with per-token asymmetric dynamic quantization, KV-cache with per-head asymmetric dynamic quantization, and weights with per-channel symmetric quantization. Calibration uses 128 samples from Wikitext-2. For each block, symmetric scaling is trained for 3 epochs, followed by weight quantization using GPTQ\cite{Frantar2022GPTQAP}, and finally, bias terms, asymmetric scaling, and learnable clipping factors are trained for 5 epochs. Learning rates are initialized as 1e-2 for scaling and clipping, 1e-3 for bias, and all are cosine-decayed throughout training. It takes 0.7 hours for 3B models and 10 hours for 70B models on a single A800 GPU.

\subsection{Main results}

\Cref{tab:main} presents the overall performance of BASE-Q and baseline methods under W4A4KV4 quantization, reporting perplexity on Wikitext-2 and the average accuracy across nine zero-shot tasks. Since BASE-Q is explicitly designed to tackle the most challenging 4-bit activation quantization, we focus solely on the W4A4KV4 configuration. Across all evaluated models, BASE-Q consistently achieves superior results, with an average accuracy drop of only 3.40\% from full precision. In comparison, the average accuracy drops for QuaRot, SpinQuant, and OSTQuant are 6.87\%, 5.95\%, and 4.80\%, respectively (averaged over eight supported models). This translates to BASE-Q narrowing the gap to full-precision by 50.5\%, 42.9\%, and 29.2\% relative to QuaRot, SpinQuant, and OSTQuant, respectively. Notably, BASE-Q performs exceptionally well on Qwen2.5-3B, where all existing methods suffer severe degradation (perplexity increases by more than 100\%). These results strongly validate our theoretical insights and demonstrate the practical effectiveness of our proposed bias correction and asymmetric scaling strategies.
\begin{table*}[!ht]
\renewcommand\arraystretch{1}
\centering
\caption{ \small Comparison on Wikitext-2 perplexity and 9 zero-shot benchmark tasks. All baseline results—QuaRot, SpinQuant, and OSTQuant—are reproduced using their official open-source implementations, with necessary modifications to support the Qwen model, which differs from LLaMA mainly by including attention bias. As the official OSTQuant repository does not support FSDP (Fully Sharded Data Parallel) training, its results are limited to models with up to 14B parameters. Complete results are provided in \Cref{appendix:result}.}
\vspace{-3.3em}
\label{tab:main}
\setlength{\tabcolsep}{1.2mm}
{\resizebox{\textwidth}{!}{
\begin{tabular}{c|cc:cc:cc|cc:cc|cc}
& & & & & & & & & & & & \\
& & & & & & & & & & & & \\
& & & & & & & & & & & & \\
& & & & & & & & & & & & \\
& & & & & & & & & & & & \\
\noalign{\vspace{0.2em}}\hline\noalign{\vspace{0.1em}}
\hline\noalign{\vspace{0.4em}}
   & \multicolumn{2}{c:}{\textbf{Qwen-2.5 3B}} & \multicolumn{2}{c:}{\textbf{Qwen-2.5 14B}} & \multicolumn{2}{c|}{\textbf{Qwen-2.5 32B}} & \multicolumn{2}{c:}{\textbf{LLaMA-3.1 8B}}  & \multicolumn{2}{c|}{\textbf{LLaMA-3.1 70B}}  & \multicolumn{2}{c:}{\textbf{LLaMA-3.2 1B}}  \\
\noalign{\vspace{0.2em}}\cdashline{2-13}\noalign{\vspace{0.3em}}
\textbf{W4A4KV4} & 0-shot$^9$ & Wiki & 0-shot$^9$ & Wiki & 0-shot$^9$ & Wiki & 0-shot$^9$ & Wiki & 0-shot$^9$ & Wiki & 0-shot$^9$ & Wiki \\
  & Avg.($\uparrow$) & ($\downarrow$) & Avg.($\uparrow$) & ($\downarrow$) & Avg.($\uparrow$) & ($\downarrow$) & Avg.($\uparrow$) & ($\downarrow$) & Avg.($\uparrow$) & ($\downarrow$) & Avg.($\uparrow$) & ($\downarrow$) \\
\noalign{\vspace{0.2em}}\hline\noalign{\vspace{0.3em}}
 Full-Precision & 64.17 & 8.03 & 70.95 & 5.29 & 71.11 & 5.02 & 68.70 & 6.23 & 73.76 & 2.81 & 55.89 & 9.75 \\
\noalign{\vspace{0.2em}}\hdashline\noalign{\vspace{0.2em}}
 QuaRot & 44.30 & 69.33 & 67.23 & 6.77 & 68.14 & 6.04 & 63.74 & 7.82 & 69.56 & 5.31 & 48.66 & 14.44 \\
 \noalign{\vspace{0.2em}}
 SpinQuant & 46.86 & 46.35 & 67.29 & 6.55 & 68.51 & 5.88 & 64.58 & 7.51 & 70.69 & 4.74 & 49.41 & 13.46 \\
 \noalign{\vspace{0.2em}}
 OSTQuant & 50.81 & 20.09 & 67.81 & 6.37 & \footnotesize{OOM} & \footnotesize{OOM} & 64.91 & 7.40 & \footnotesize{OOM} & \footnotesize{OOM} & \textbf{50.85} & 12.84 \\
 \noalign{\vspace{0.2em}}
\rowcolor{yellow!20} \textbf{BASE-Q} & \textbf{58.13} & \textbf{10.83} & \textbf{68.90} & \textbf{6.28} & \textbf{70.18} & \textbf{5.65} & \textbf{65.36} & \textbf{7.17} & \textbf{71.54} & \textbf{4.17} & 50.61 & \textbf{12.66}  \\
\noalign{\vspace{0.2em}}\hline\noalign{\vspace{0.1em}}
\hline\noalign{\vspace{0.4em}}
  & \multicolumn{2}{c:}{\textbf{Llama-2 7B}} & \multicolumn{2}{c:}{\textbf{Llama-2 13B}} & \multicolumn{2}{c|}{\textbf{Llama-2 70B}} & \multicolumn{2}{c:}{\textbf{LLaMA-3 8B}}  & \multicolumn{2}{c|}{\textbf{LLaMA-3 70B}}  & \multicolumn{2}{c}{\textbf{LLaMA-3.2 3B}}\\
\noalign{\vspace{0.2em}}\cdashline{2-13}\noalign{\vspace{0.3em}}
\textbf{W4A4KV4} & 0-shot$^9$ & Wiki & 0-shot$^9$ & Wiki & 0-shot$^9$ & Wiki & 0-shot$^9$ & Wiki & 0-shot$^9$ & Wiki & 0-shot$^9$ & Wiki \\
  & Avg.($\uparrow$) & ($\downarrow$) & Avg.($\uparrow$) & ($\downarrow$) & Avg.($\uparrow$) & ($\downarrow$) & Avg.($\uparrow$) & ($\downarrow$) & Avg.($\uparrow$) & ($\downarrow$) & Avg.($\uparrow$) & ($\downarrow$) \\
\noalign{\vspace{0.2em}}\hline\noalign{\vspace{0.3em}}
  Full-Precision & 65.22 & 5.47 & 67.62 & 4.88 & 71.57 & 3.32 & 68.11 & 6.14 & 73.82 & 2.86 & 63.59 & 7.81 \\
\noalign{\vspace{0.2em}}\hdashline\noalign{\vspace{0.2em}}
 QuaRot & 61.59 & 6.12 & 65.03 & 5.39 & 70.28 & 3.76 & 62.89 & 7.82 & 68.76 & 5.62 & 55.86 & 10.07 \\
 \noalign{\vspace{0.2em}}
 SpinQuant & 61.38 & 5.99 & 65.63 & 5.30 & 70.22 & 3.71 & 63.80 & 7.49 & 69.93 & 5.11 & 57.74 & 9.32\\
\noalign{\vspace{0.2em}}
 OSTQuant & 62.08 & 5.92 & 65.43 & 5.24 & \footnotesize{OOM} & \footnotesize{OOM} & 64.72 & 7.36 & \footnotesize{OOM} & \footnotesize{OOM}  & 59.29 & 9.16 \\
 \noalign{\vspace{0.2em}}
\rowcolor{yellow!20} \textbf{BASE-Q} & \textbf{62.50} & \textbf{5.85} & \textbf{65.95} & \textbf{5.19} & \textbf{70.74} & \textbf{3.59} & \textbf{65.60} & \textbf{7.12} & \textbf{71.83} & \textbf{4.06}& \textbf{60.02} & \textbf{9.01} \\
\noalign{\vspace{0.2em}}\hline\noalign{\vspace{0.1em}}
\hline
\end{tabular}}}
\vspace{-1em}
\end{table*}

\subsection{Prefill acceleration on GPU}

To evaluate real-world acceleration, we benchmark INT4 quantization during the compute-bound prefill stage on an NVIDIA 3090 GPU. All experiments use a sequence length of 2048, with batch sizes varying from 1 to 64. 4-bit matrix multiplications are implemented by the NVIDIA's Cutlass library, while other custom operators are written in Triton for flexibility and speed. To minimize overhead, we fuse the online bias and scaling introduced by BASE-Q within quantization and de-quantization into a single Triton kernel (\Cref{fig:fig6}), rendering the additional costs of bias and scaling computation negligible.
For the online Hadamard transform, we find that the fast algorithm~\cite{fast-hadamard-transform} adopted in Quarot, which uses the recursive Cooley-Tukey method to achieve $O(nlogn)$ complexity, does not fully leverage tensor core parallelism on modern GPUs. To address this, we implement the Hadamard transform as a larger matrix multiplication in Triton, better utilizing available compute. For query and key projections (with small dimensions, e.g., 128×128), we perform the transformation directly as a matrix multiplication. For the much larger \textit{down} layer, we only apply one single Cooley-Tukey step, ensuring all compute-intensive kernels operate on manageable, high-parallelism sub-matrices.
As shown in \Cref{fig:fig5}, our approach yields $2.1\times$ to $2.4\times$ acceleration across all batch sizes compared with standard FP16. Notably, our optimizations incur only minimal overhead versus pure INT4 quantization without any online operations.

\begin{minipage}{0.49\linewidth}
  \centering
  \includegraphics[width=\linewidth]{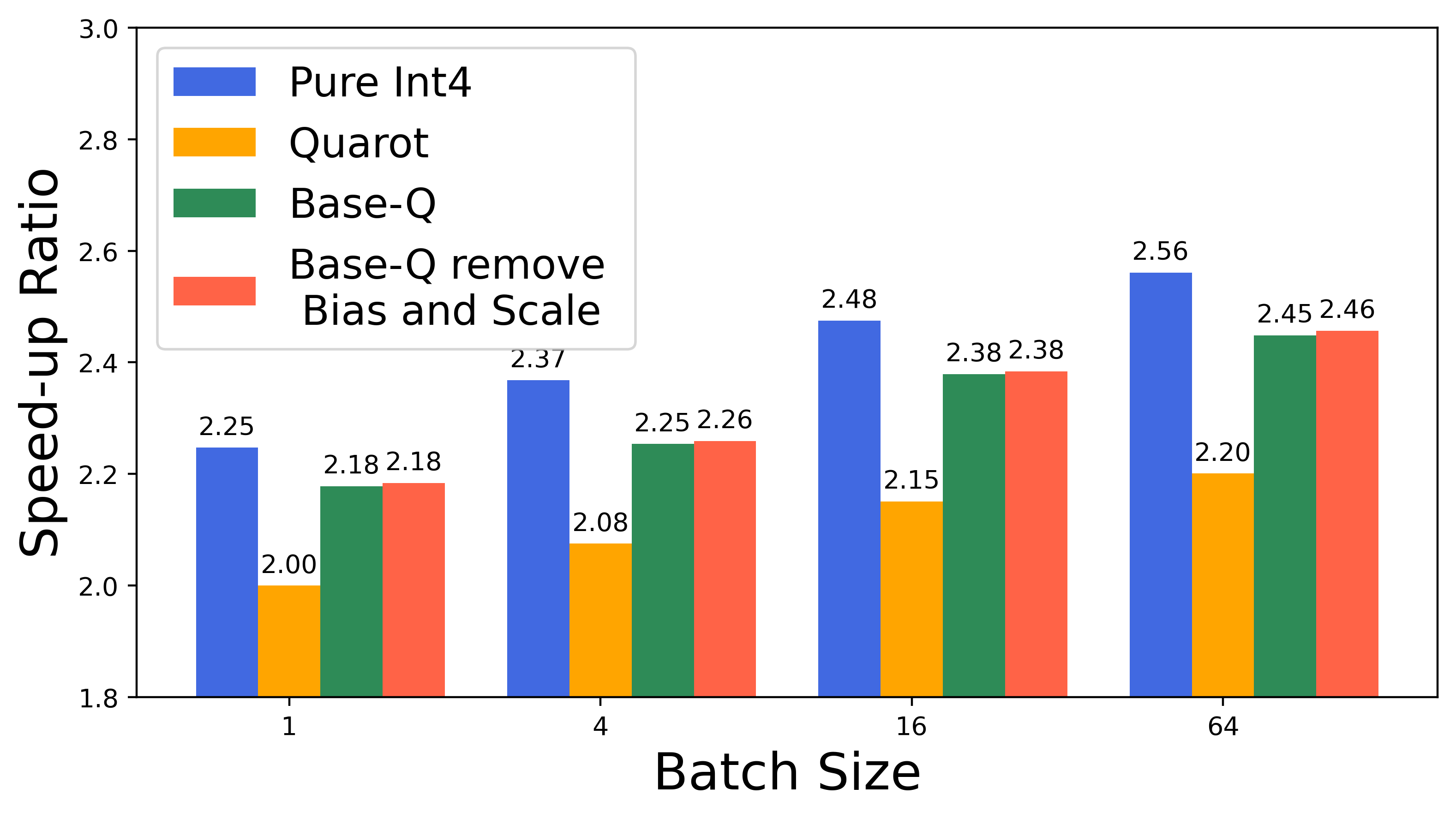}
  \captionof{figure}{Prefill speedup for Llama2-7B with seqlens 2048 as batch size scales from 1 to 64.}
  \label{fig:fig5}
\end{minipage}
\hfill
\begin{minipage}{0.48\linewidth}
  \centering
  \includegraphics[width=0.9\linewidth]{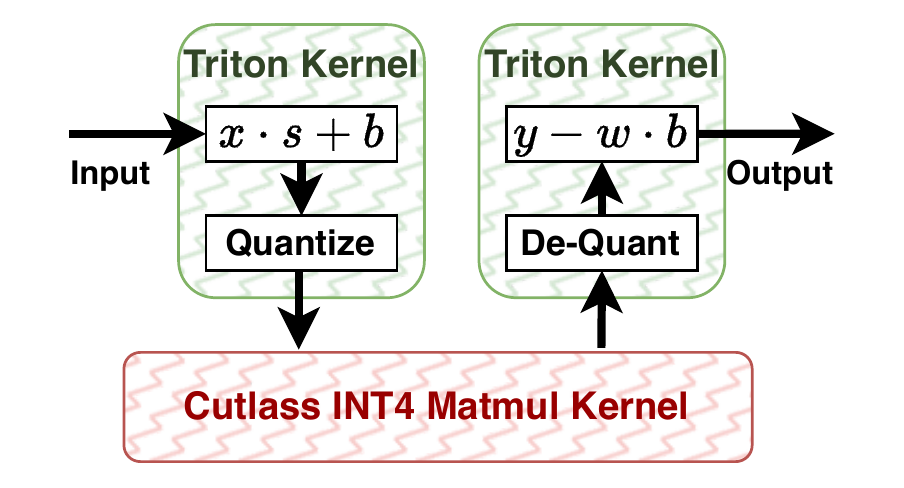}
  \captionof{figure}{Illustration of kernel fusion.}
  \label{fig:fig6}
\end{minipage}

\subsection{Ablation study}

We perform a systematic ablation study on Qwen2.5-3B, Llama2-7B, and Llama3-8B to assess the effects of different quantization strategies in BASE-Q, as well as those from Quarot, SpinQuant, and OSTQuant. Our experiments as shown in \Cref{tab:ablation} indicate that bias correction yields substantial perplexity reductions for Qwen2.5-3B and Llama3-8B, highlighting the importance of addressing inter-channel bias in these models. In contrast, its effect on Llama2-7B is marginal. Notably, asymmetric scaling delivers consistent improvements across all three models, confirming its effectiveness in mitigating the adverse impact of clipping.

\definecolor{deepgreen}{RGB}{0,100,0}
\definecolor{deepred}{RGB}{139,0,0}
\begin{table*}[!htb]
\renewcommand\arraystretch{1}
\centering
\caption{ \small Ablation Study on WikiText2 (word perplexity$\textcolor{deepgreen}{\downarrow}$)
}
\vspace{-1.3em}
\label{tab:ablation}
\setlength{\tabcolsep}{0.8mm}
{\resizebox{0.85\textwidth}{!}{
\begin{tabular}{c|c:c:c:c:c|c:c:c}
& & & & & & &  \\
\hline\noalign{\vspace{0.1em}}
 Method & \makecell{\small{Fixed}\\\small{Rotation}} & \makecell{\small{Learned}\\\small{Rotation}} & \makecell{\small{Bias}\\\small{Corect.}}  & \makecell{\small{Unpaired}\\\small{Scale}} & Scale  & \small{Qwen2.5-3B} & \small{Llama-2-7B} &  \small{Llama-3-8B} \\
\noalign{\vspace{0.1em}}\hline\noalign{\vspace{0.1em}}
 QuaRot & \small{$\textcolor{brown}{\boldsymbol{\mathit{R}}_{res}}\;\boldsymbol{\mathit{R}}_{v}\;\boldsymbol{\mathit{R}}_{qk}\;\boldsymbol{\mathit{R}}_{down}$} &  &  &  &  & 69.33 & 6.12& 7.82 \\
 \noalign{\vspace{0.1em}}\hdashline\noalign{\vspace{0.1em}}
  & $\textcolor{brown}{\boldsymbol{\mathit{R}}_{res}}\;\boldsymbol{\mathit{R}}_{qk}\;\boldsymbol{\mathit{R}}_{down}$ & $\boldsymbol{\mathit{R}}_{v}$   &  &  &  & 54.38 \scriptsize{\textcolor{deepgreen}{-5.95}}  & 6.18 \scriptsize{\textcolor{deepred}{+0.06}} & 7.63 \scriptsize{\textcolor{deepgreen}{-0.19}}\\
 \noalign{\vspace{0.1em}}\noalign{\vspace{0.1em}}
  & $\textcolor{brown}{\boldsymbol{\mathit{R}}_{res}}\;\boldsymbol{\mathit{R}}_{qk}\;\boldsymbol{\mathit{R}}_{down}$ & $\boldsymbol{\mathit{R}}_{v}$   & \ding{51} & &    & 13.59 \scriptsize{\textcolor{deepgreen}{-40.79}} & 6.12 \scriptsize{\textcolor{deepgreen}{-0.06}}& 7.42 \scriptsize{\textcolor{deepgreen}{-0.21}}\\
 \noalign{\vspace{0.1em}}\noalign{\vspace{0.1em}}
   & $\textcolor{brown}{\boldsymbol{\mathit{R}}_{res}}\;\boldsymbol{\mathit{R}}_{qk}\;\boldsymbol{\mathit{R}}_{down}$ & $\boldsymbol{\mathit{R}}_{v}$  & \ding{51} & \ding{51} &  & 10.82 \scriptsize{\textcolor{deepgreen}{-2.77}} & 5.92 \scriptsize{\textcolor{deepgreen}{-0.20}} & 7.22 \scriptsize{\textcolor{deepgreen}{-0.20}}\\
 \noalign{\vspace{0.1em}}\noalign{\vspace{0.1em}}
 \textbf{BASE-Q} & $\textcolor{brown}{\boldsymbol{\mathit{R}}_{res}}\;\boldsymbol{\mathit{R}}_{qk}\;\boldsymbol{\mathit{R}}_{down}$ & $\boldsymbol{\mathit{R}}_{v}$ & \ding{51} & \ding{51} & \ding{51} & 10.83 \scriptsize{\textcolor{deepred}{+0.01}}  & 5.85 \scriptsize{\textcolor{deepgreen}{-0.07}}& 7.14 \scriptsize{\textcolor{deepgreen}{-0.08}}\\
 \noalign{\vspace{0.1em}}\hdashline\noalign{\vspace{0.1em}}
 SpinQuant & $\boldsymbol{\mathit{R}}_{qk}\;\boldsymbol{\mathit{R}}_{down}$ &  $\textcolor{brown}{\boldsymbol{\mathit{R}}_{res}}\;\boldsymbol{\mathit{R}}_{v}$  &  &  &  & 46.35 & 5.99 & 7.49\\
 \noalign{\vspace{0.1em}}\noalign{\vspace{0.1em}}
 OSTQuant & $\boldsymbol{\mathit{R}}_{qk}\;\boldsymbol{\mathit{R}}_{down}$ &  ${\textcolor{brown}{\boldsymbol{\mathit{R}}_{res}}}\;\boldsymbol{\mathit{R}}_{v}$  & &  &\ding{51} & 20.09 & 5.92  & 7.36\\
\noalign{\vspace{0.7em}}\hline
\end{tabular}}}
\vspace{-1em}
\end{table*}

We further performed ablation studies where, within the BASE-Q framework, global rotations were respectively implemented as a standard Hadamard matrix, a random Hadamard matrix, and a learnable rotation (following SpinQuant’s official codebase). In SpinQuant (cf. Fig. 4 in \cite{Liu2024SpinQuantLQ}), the authors observed that using random Hadamard rotations could cause large fluctuations and subpar quantization performance. Based on this observation, SpinQuant advocates for learnable global rotations to stabilize and improve quantization results under their framework.

However, in our ablation studies \Cref{tab:ablation2} we found that, across five benchmark LLMs, the quantization metrics showed negligible differences across different rotation types within our BASE-Q framework. This suggests that any potential improvements from learnable global rotations are effectively subsumed by BASE-Q’s bias correction component. Therefore, the additional memory and computational cost introduced by learnable global rotations is unnecessary for maintaining quantization quality in our method.

\begin{table*}[!ht]
\renewcommand\arraystretch{1}
\centering
\caption{ \small Ablation study on $R_{res}$ choice.}
\vspace{-1.8em}
\label{tab:ablation2}
\setlength{\tabcolsep}{1.2mm}
{\resizebox{\textwidth}{!}{
\begin{tabular}{cc|cc:cc|cc:cc:cc}
& & & & & & & & & & &   \\
& & & & & & & & & & &   \\
\noalign{\vspace{0.2em}}\hline\noalign{\vspace{0.1em}}
\hline\noalign{\vspace{0.4em}}
   & & \multicolumn{2}{c:}{\textbf{Qwen-2.5 3B}} & \multicolumn{2}{c:}{\textbf{Qwen-2.5 14B}} & \multicolumn{2}{c|}{\textbf{LLaMA-2 7B}} & \multicolumn{2}{c:}{\textbf{LLaMA-3 8B}}  & \multicolumn{2}{c|}{\textbf{LLaMA-2 13B}}   \\
\noalign{\vspace{0.2em}}\cdashline{2-11}\noalign{\vspace{0.3em}}
\textbf{W4A4KV4} & \textbf{$R_{res}$ Setting} & 0-shot$^9$ & Wiki & 0-shot$^9$ & Wiki & 0-shot$^9$ & Wiki & 0-shot$^9$ & Wiki & 0-shot$^9$ & Wiki \\
  & & Avg.($\uparrow$) & ($\downarrow$) & Avg.($\uparrow$) & ($\downarrow$) & Avg.($\uparrow$) & ($\downarrow$) & Avg.($\uparrow$) & ($\downarrow$) & Avg.($\uparrow$) & ($\downarrow$)\\
\noalign{\vspace{0.2em}}\hline\noalign{\vspace{0.3em}}
 Full-Precision && 64.17 & 8.03 & 70.95 & 5.29 & 65.22 & 5.47 & 68.11 & 6.14 & 67.62 & 4.88 \\
\noalign{\vspace{0.2em}}\hdashline\noalign{\vspace{0.2em}}
 QuaRot & Random Hadamard & 44.30 & 69.33 & 67.23 & 6.77 & 61.59 & 6.12 & 62.89 & 7.82 & 65.03 & 5.39\\
 \noalign{\vspace{0.2em}}
 SpinQuant & Learned Rotation & 46.86 & 46.35 & 67.29 & 6.55 & 61.38 & 5.99 & 63.80 & 7.49 & 65.63 & 5.30 \\
 \noalign{\vspace{0.2em}}
\textbf{BASE-Q} & Standard Hadamard& \textbf{58.93} & \textbf{10.43} & \textbf{69.17} & \textbf{6.28} & \textbf{62.08} & \textbf{5.85} & \textbf{65.33} & \textbf{7.13} & \textbf{65.89} & \textbf{5.20}  \\
& Random Hadamard& \textbf{58.60} & \textbf{10.43} & \textbf{69.22} & \textbf{6.28} & \textbf{62.24} & \textbf{5.87} & \textbf{65.21} & \textbf{7.14} & \textbf{65.46} & \textbf{5.20}  \\
& Learned Hadamard& \textbf{57.95} & \textbf{10.44} & \textbf{68.43} & \textbf{6.26} & \textbf{62.68} & \textbf{5.86} & \textbf{65.01} & \textbf{7.13} & \textbf{65.60} & \textbf{5.20}  \\
\noalign{\vspace{0.2em}}\hline\noalign{\vspace{0.1em}}
\hline
\end{tabular}}}
\vspace{-1em}
\end{table*}

\subsection{Resource and accuracy trade-offs}
\label{sec:discussion}

Recent studies such as FlatQuant\cite{sun2024flatquant} employ layer-wise learned online matrix transforms to improve the performance of quantized linear layers, at the cost of increased inference overhead. In contrast, BASE-Q applies rotations $\boldsymbol{\mathit{R}}_{res}$ that can be fused into the weights, eliminating the need for the extra computation. As shown in \Cref{fig:pareto}, BASE-Q consistently achieves \textbf{Pareto-optimal} trade-offs between accuracy and inference overhead, outperforming Quarot at comparable cost, while FlatQuant suffers from significant computation overhead. Furthermore, FlatQuant suffers from significant instability on Qwen2.5-3B, where the quantized model collapses. \Cref{fig:memory} further demonstrates the lower GPU memory usage of BASE-Q under quantization process. Importantly, our method is \textbf{fully orthogonal} to other advanced methods and can flexibly combine with them, allowing practitioners to efficiently adapt to different deployment and training requirements.

\begin{minipage}{0.47\linewidth}
  \centering
  \includegraphics[width=0.95\linewidth]{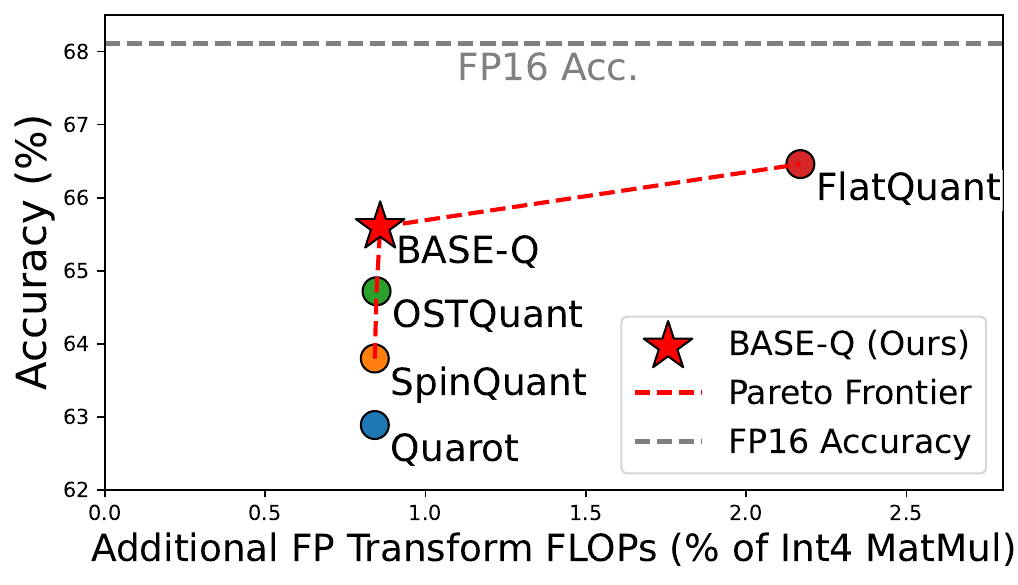}
  \captionof{figure}{Zero-shot$^9$ accuracy and inference overhead on Llama3-8B. BASE-Q achieves Pareto-optimal trade-offs. Note that hadamard rotations require only addition operations compared to standard matrix transformation.}
  \label{fig:pareto}
\end{minipage}
\hfill
\begin{minipage}{0.47\linewidth}
  \centering
  \includegraphics[width=0.95\linewidth]{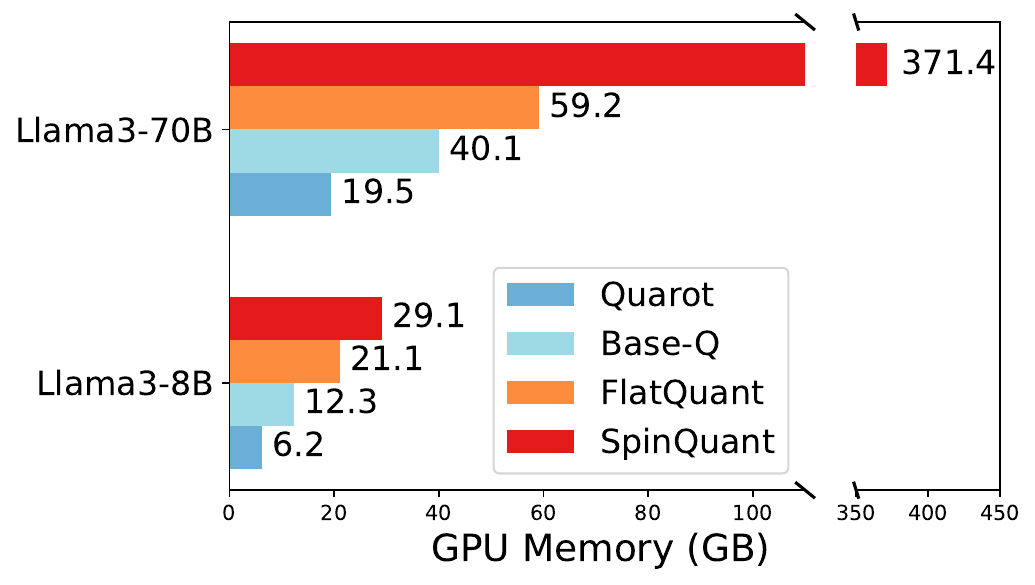}
  \captionof{figure}{GPU memory consumption during optimization. BASE-Q enables quantizing a 70B model on a single A800 GPU within 10 hours, whereas SpinQuant requires at least 5 A800 GPUs and a total of 36 GPU-hours.}
  \label{fig:memory}
\end{minipage}

\section{Conclusion}
In this work, we analyze key challenges in rotational quantization for large language models, identifying channel mean discrepancies and cumulative clipping-induced energy loss as major sources of quantization errors. To address these issues, we propose BASE-Q, which introduces blockwise bias correction and per-channel asymmetric scaling to achieve accurate and efficient quantization with minimal resource overhead. Experiments on popular LLM benchmarks demonstrate that BASE-Q closes the accuracy gap to floating-point baselines while substantially reducing memory usage, enabling single-GPU quantization even for large-scale models. Our results highlight BASE-Q as a practical and scalable approach for quantizing large language models.

\bibliographystyle{unsrt}   
\bibliography{reference}   

\appendix
\newpage
\section{Full quantization results}
\label{appendix:result}
We present comprehensive quantization results in this section, including perplexity on WikiText2 and zero-shot accuracy on nine evaluation datasets. All baseline results, including QuaRot, SpinQuant, and OSTQuant, are reproduced using their official open-source implementations, with necessary modifications to support the Qwen model, which differs from LLaMA mainly by including attention bias. As the official OSTQuant repository does not support FSDP (Fully Sharded Data Parallel) training, its results are limited to models with up to 14B parameters, while larger models, such as the 70B model, encounter out-of-memory (OOM) issues. Results for the Llama2 series are summarized in \Cref{tab:llama2_comparison}, those for the Llama3 series are in \Cref{tab:llama3_comparison}, and for the Qwen2.5 series in \Cref{tab:qwen25_comparison}.

\input{tables/appendix_all_results}
\section{Visualization of activation distribution}
\label{appendix:visual}
We provide visualization results for the input activations of the multi-head attention blocks and MLP blocks at the 1st, 11th, and 31st layers of Qwen2.5-3B, Llama2-7B, and Llama3-8B. For each selected layer, we illustrate the distributions of activations under various rotation strategies. We observe that inter-channel mean misalignment consistently occurs in different layers and across all evaluated models, and that learned rotations are insufficient to eliminate this issue. As discussed in \Cref{sec:round-error}, this misalignment is a significant source of quantization error, which can be effectively mitigated by our proposed bias correction technique.
\begin{figure} [h]
	\begin{center}
	\includegraphics[width=1\textwidth]{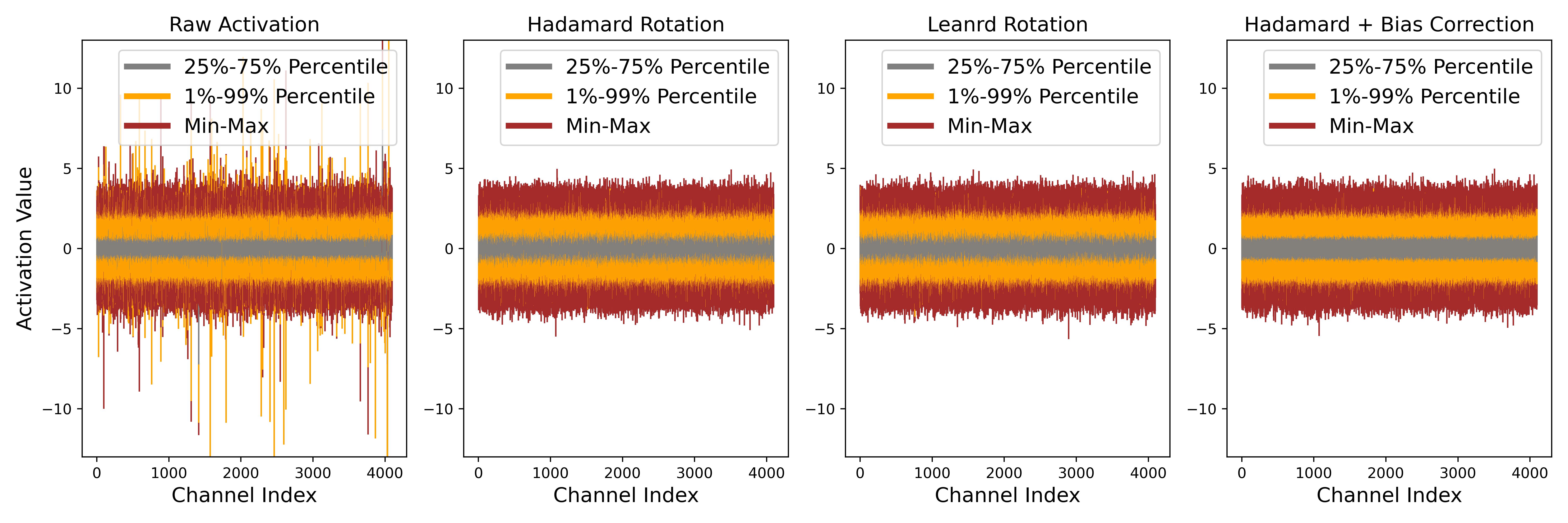}
	\caption{\small Visualizations comparing the activation distributions from the 1st MHSA block in Llama2-7B.
}
        \end{center}
\end{figure} 

\begin{figure} [h]
	\begin{center}
	\includegraphics[width=1\textwidth]{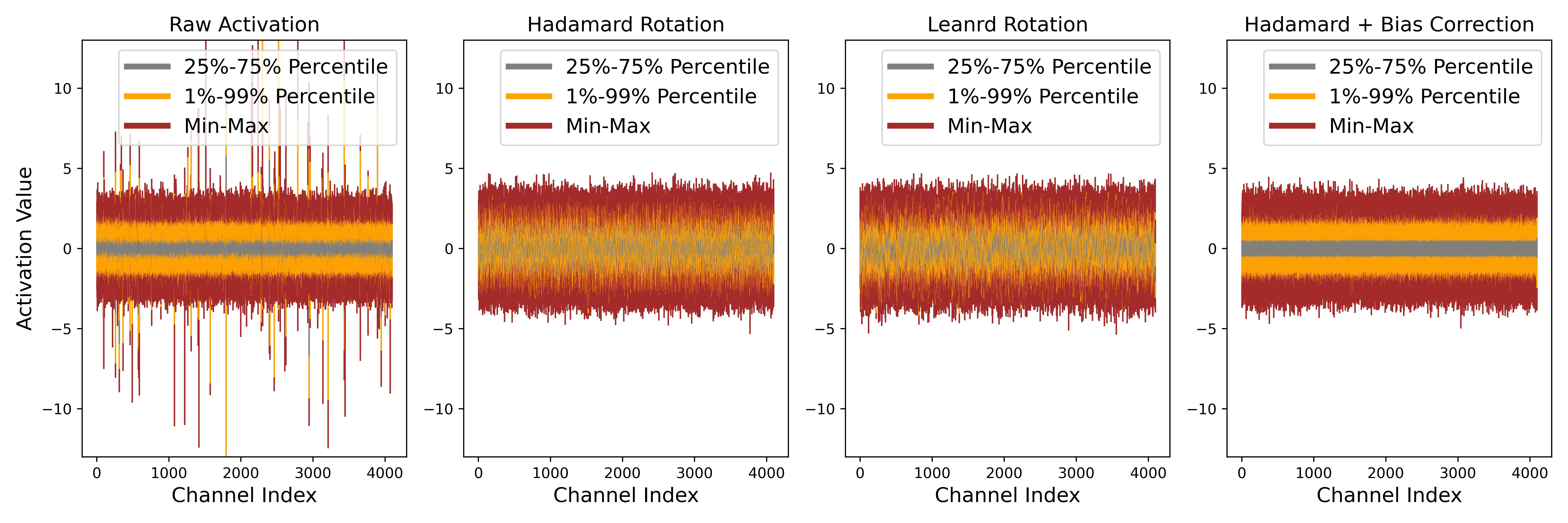}
	\caption{\small Visualizations comparing the activation distributions from the 1st MLP block in Llama2-7B.
}
        \end{center}
\end{figure} 
\begin{figure} [h]
	\begin{center}
	\includegraphics[width=1\textwidth]{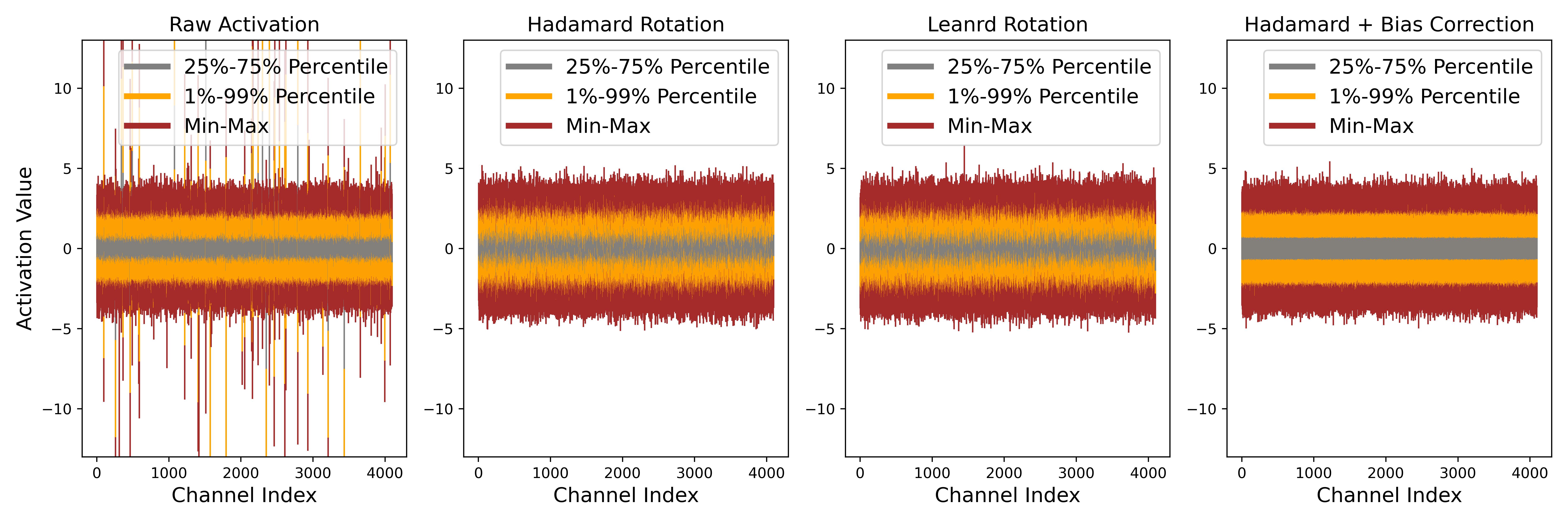}
	\caption{\small Visualizations comparing the activation distributions from the 11th MHSA block in Llama2-7B.
}
        \end{center}
\end{figure} 

\begin{figure} [h]
	\begin{center}
	\includegraphics[width=1\textwidth]{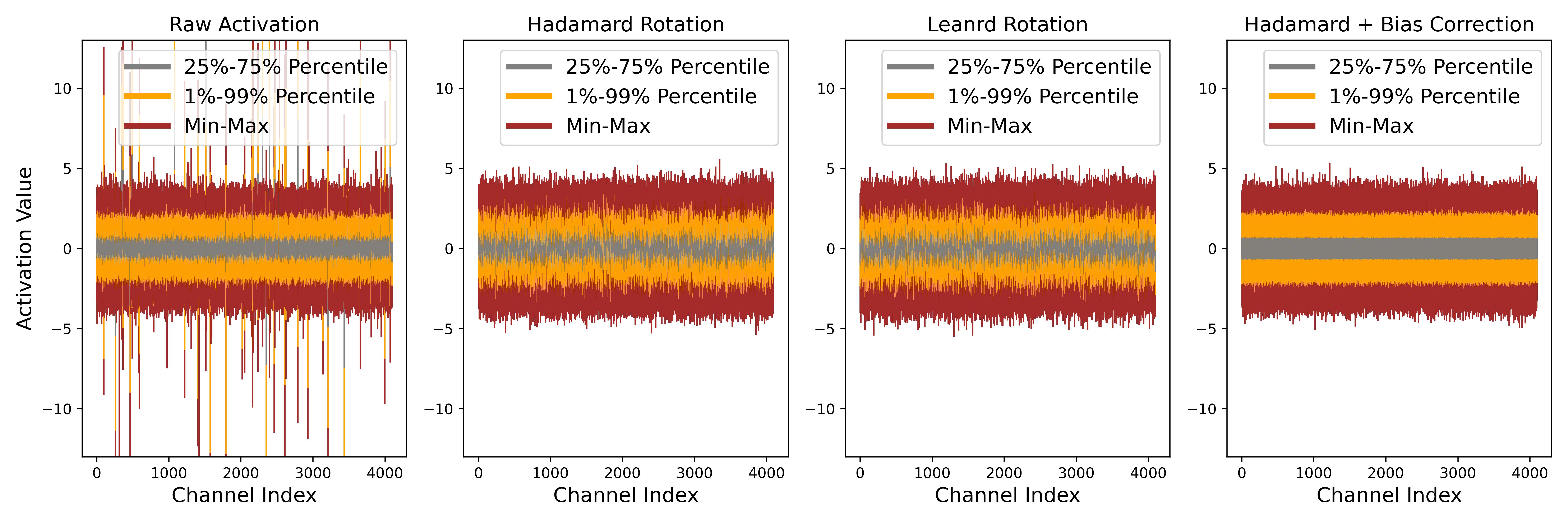}
	\caption{\small Visualizations comparing the activation distributions from the 11th MLP block in Llama2-7B.
}
        \end{center}
\end{figure} 

\begin{figure} [h]
	\begin{center}
	\includegraphics[width=1\textwidth]{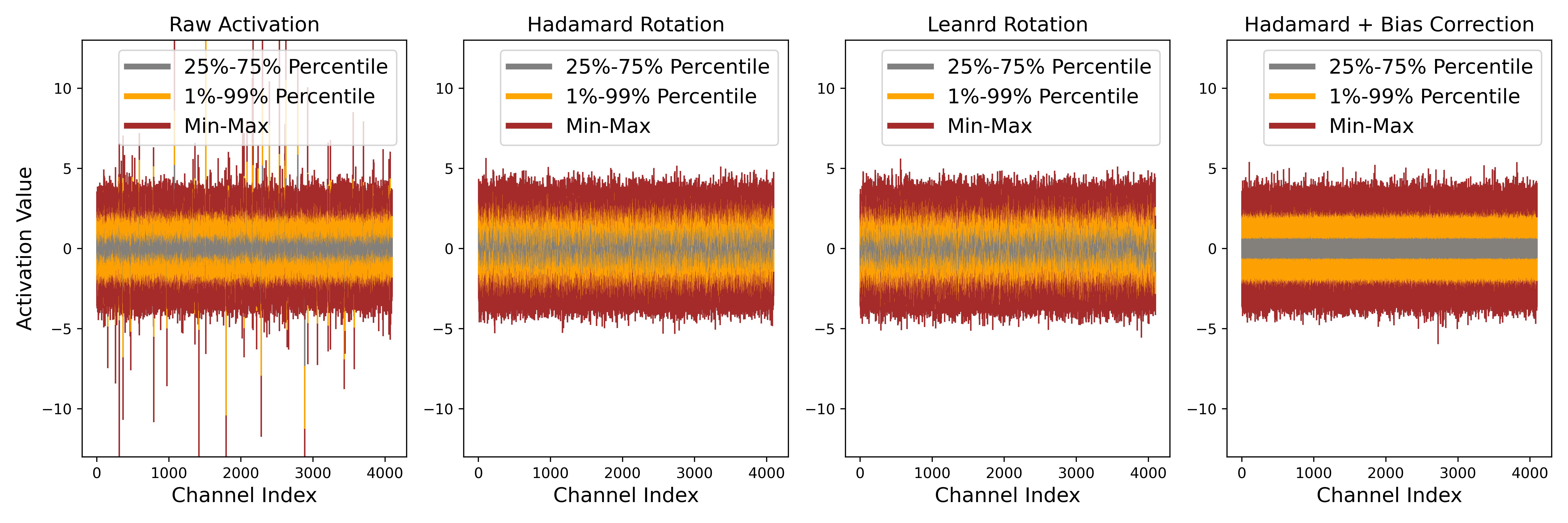}
	\caption{\small Visualizations comparing the activation distributions from the 31st MHSA block in Llama2-7B.
}
        \end{center}
\end{figure} 

\begin{figure} [h]
	\begin{center}
	\includegraphics[width=1\textwidth]{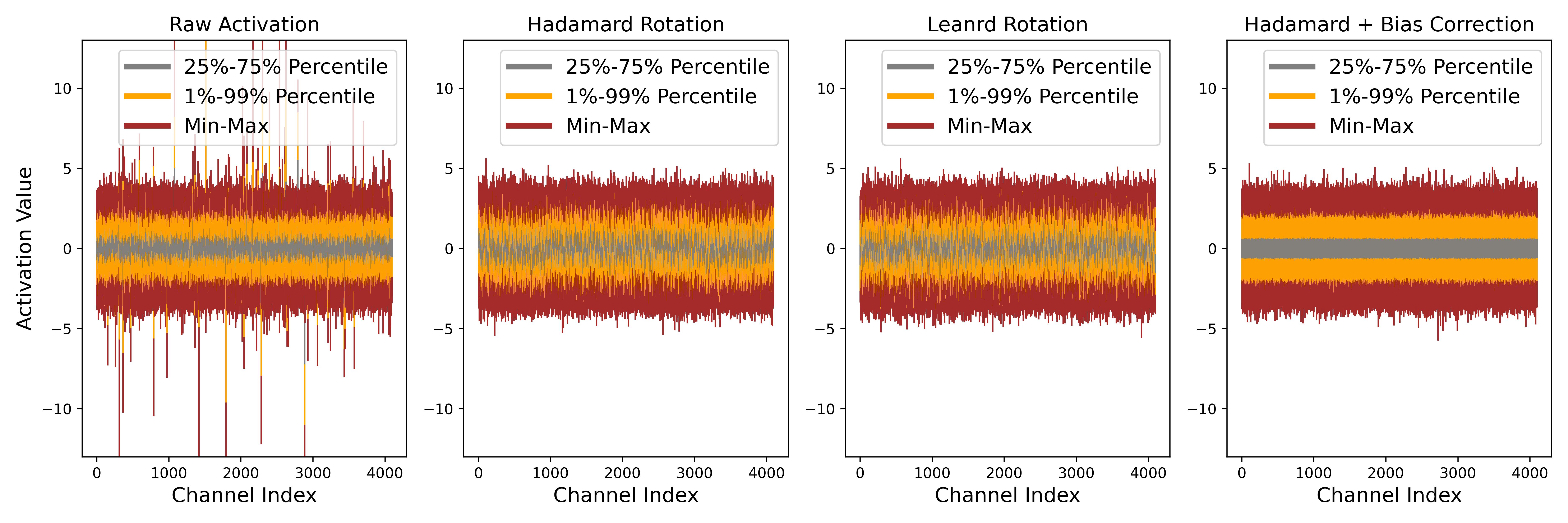}
	\caption{\small Visualizations comparing the activation distributions from the 31st MLP block in Llama2-7B.
}
        \end{center}
\end{figure} 

\begin{figure} [h]
	\begin{center}
	\includegraphics[width=1\textwidth]{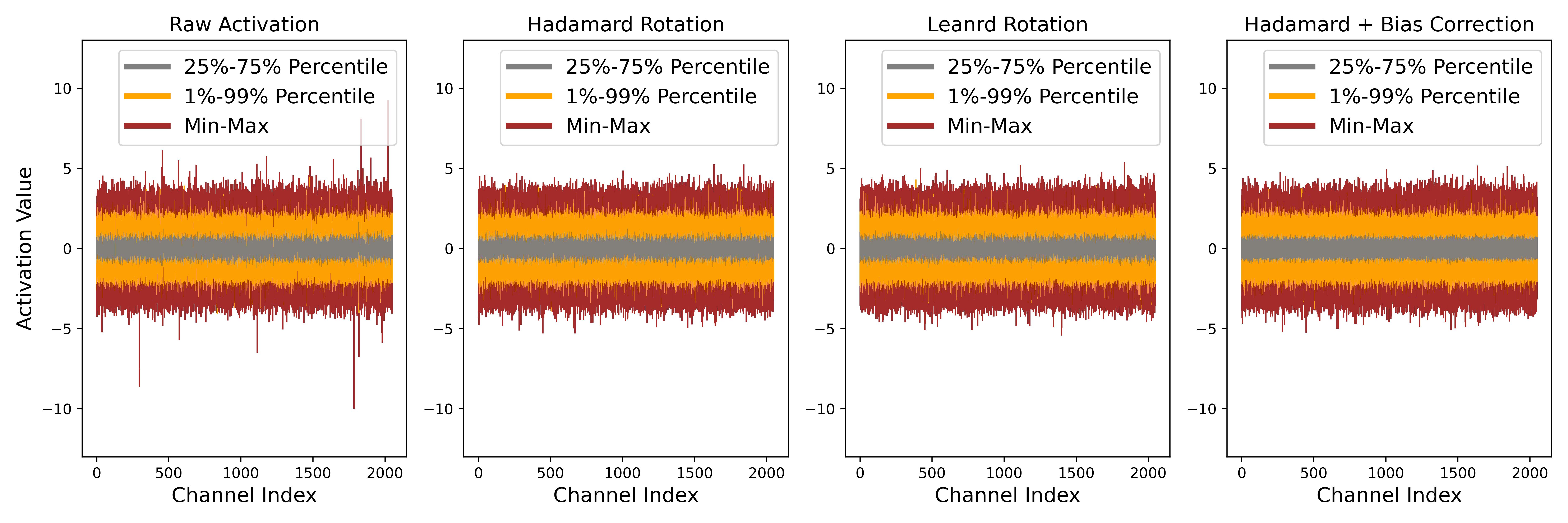}
	\caption{\small Visualizations comparing the activation distributions from the 1st MHSA block in Qwen2.5-3B.
}
        \end{center}
\end{figure} 

\begin{figure} [h]
	\begin{center}
	\includegraphics[width=1\textwidth]{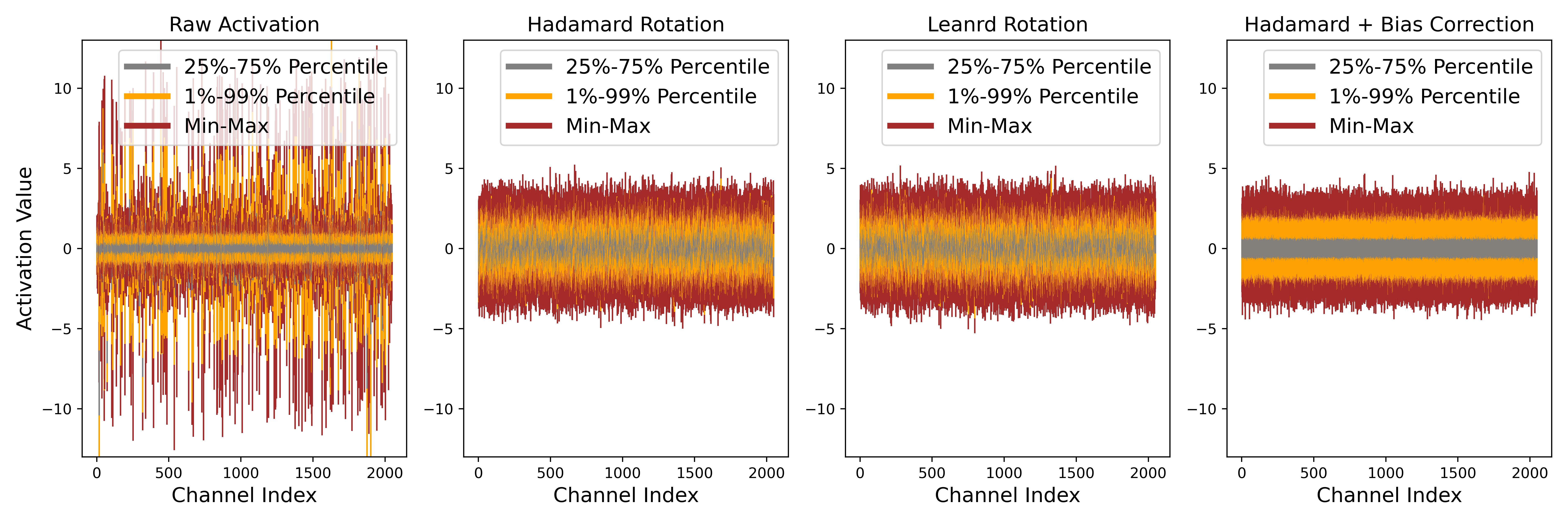}
	\caption{\small Visualizations comparing the activation distributions from the 1st MLP block in Qwen2.5-3B.
}
        \end{center}
\end{figure} 

\begin{figure} [h]
	\begin{center}
	\includegraphics[width=1\textwidth]{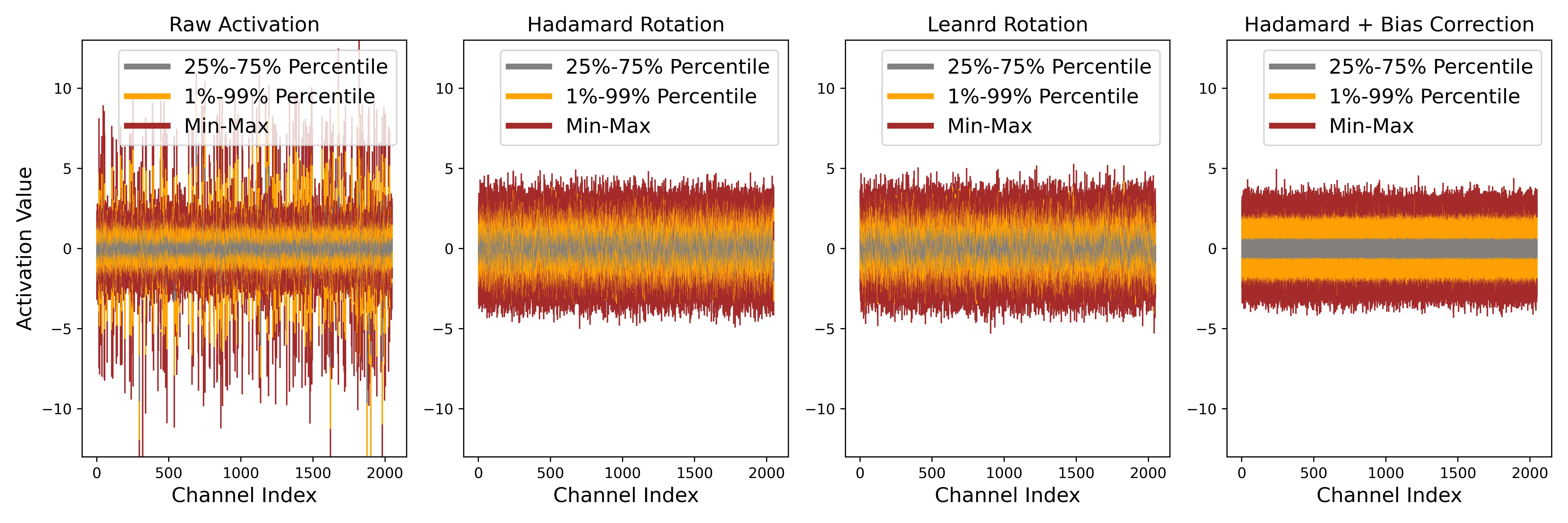}
	\caption{\small Visualizations comparing the activation distributions from the 11th MHSA block in Qwen2.5-3B.
}
        \end{center}
\end{figure} 

\begin{figure} [h]
	\begin{center}
	\includegraphics[width=1\textwidth]{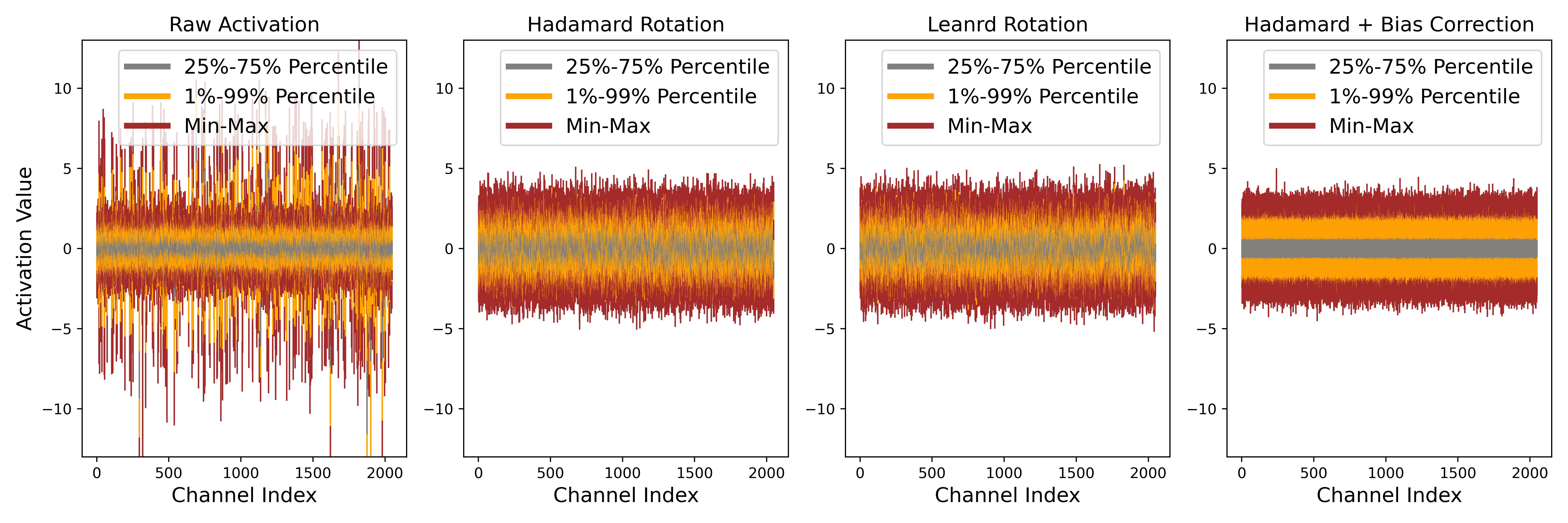}
	\caption{\small Visualizations comparing the activation distributions from the 11th MLP block in Qwen2.5-3B.
}
        \end{center}
\end{figure} 

\begin{figure} [h]
	\begin{center}
	\includegraphics[width=1\textwidth]{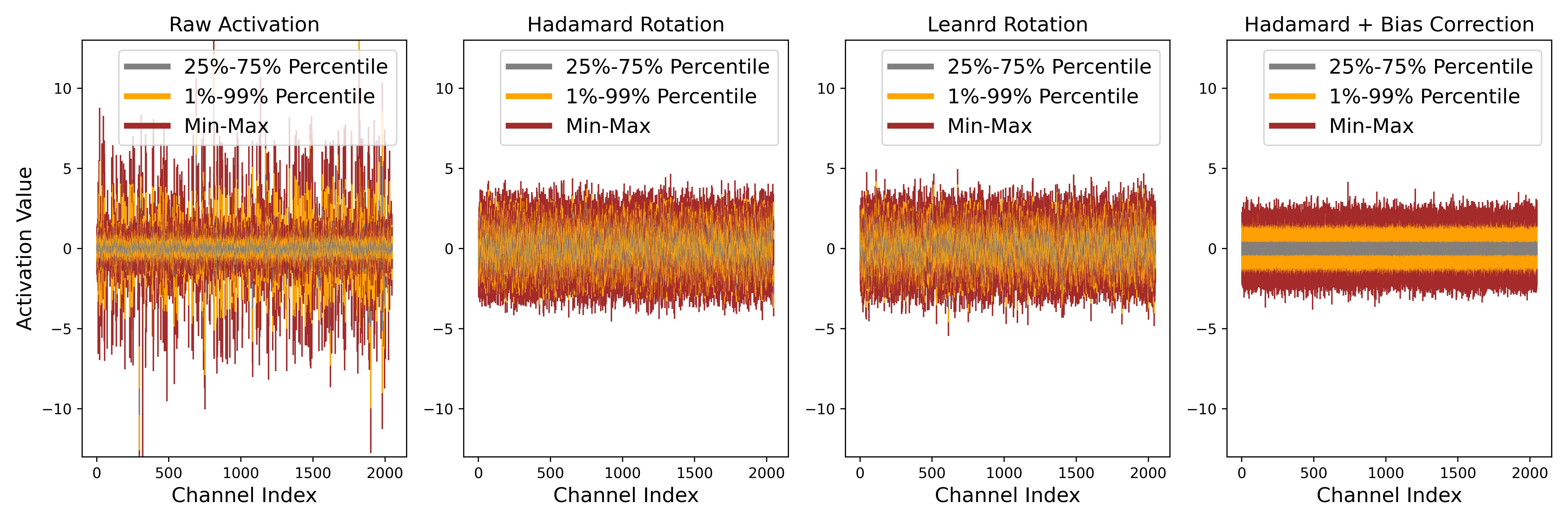}
	\caption{\small Visualizations comparing the activation distributions from the 31st MHSA block in Qwen2.5-3B.
}
        \end{center}
\end{figure} 

\begin{figure} [h]
	\begin{center}
	\includegraphics[width=1\textwidth]{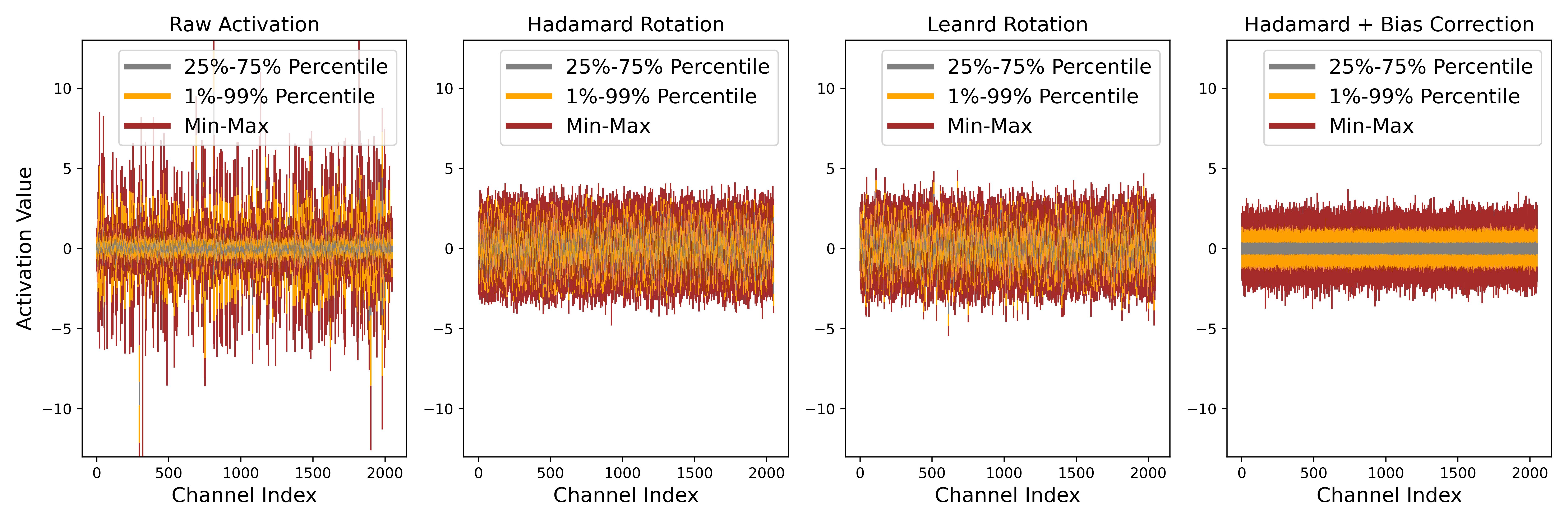}
	\caption{\small Visualizations comparing the activation distributions from the 31st MLP block in Qwen2.5-3B.
}
        \end{center}
\end{figure} 

\begin{figure} [h]
	\begin{center}
	\includegraphics[width=1\textwidth]{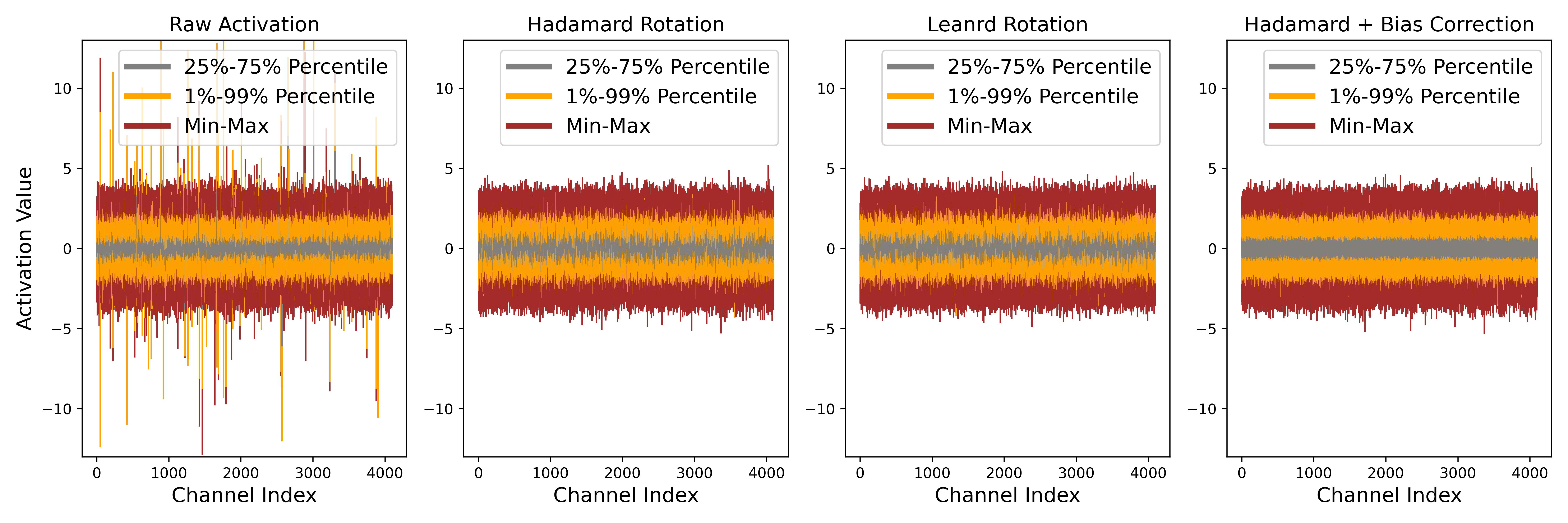}
	\caption{\small Visualizations comparing the activation distributions from the 1st MHSA block in Llama3-8B.
}
        \end{center}
\end{figure} 

\begin{figure} [h]
	\begin{center}
	\includegraphics[width=1\textwidth]{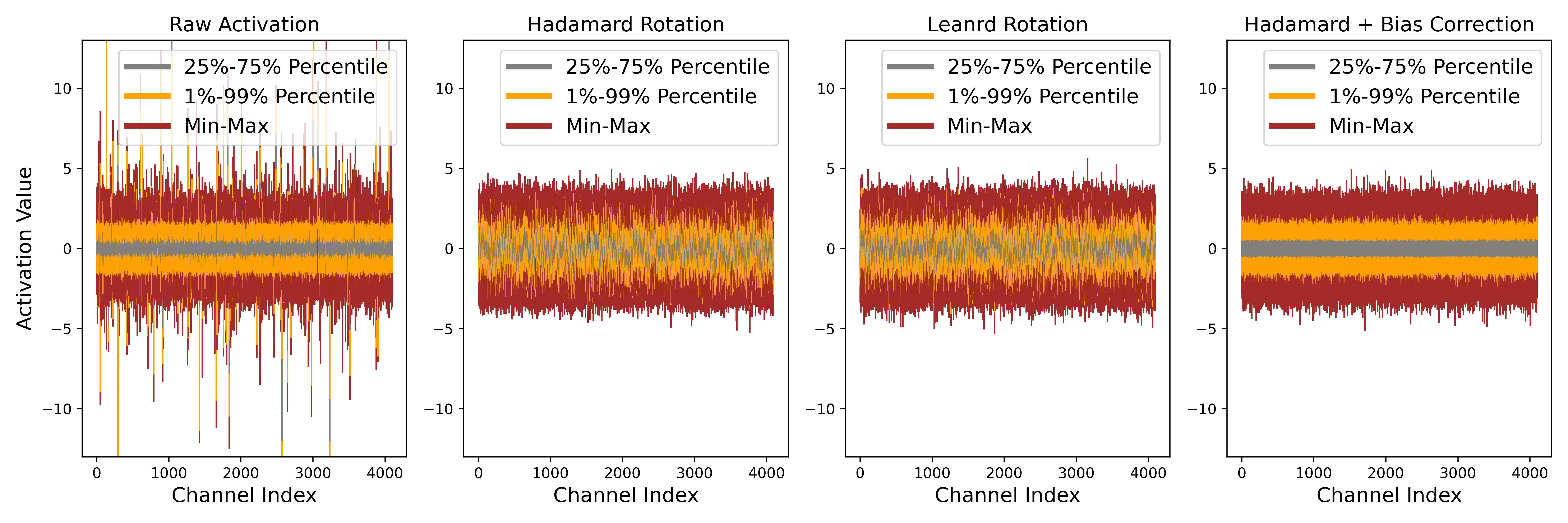}
	\caption{\small Visualizations comparing the activation distributions from the 1st MLP block in Llama3-8B.
}
        \end{center}
\end{figure} 

\begin{figure} [h]
	\begin{center}
	\includegraphics[width=1\textwidth]{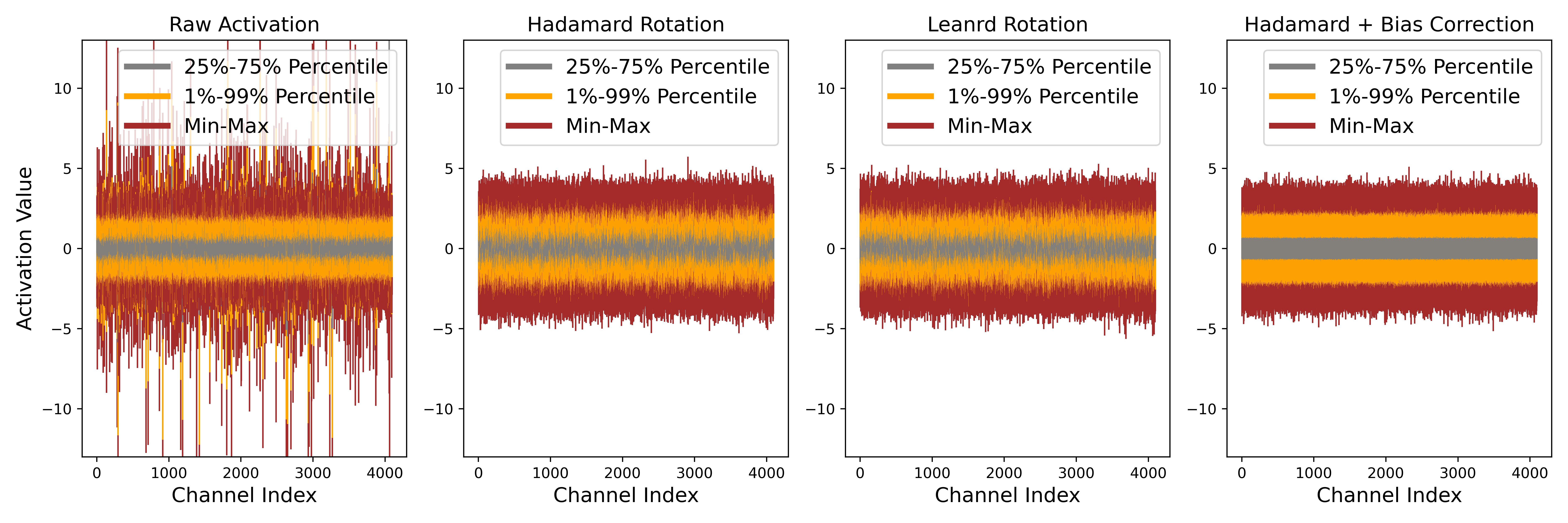}
	\caption{\small Visualizations comparing the activation distributions from the 11th MHSA block in Llama3-8B.
}
        \end{center}
\end{figure} 

\begin{figure} [h]
	\begin{center}
	\includegraphics[width=1\textwidth]{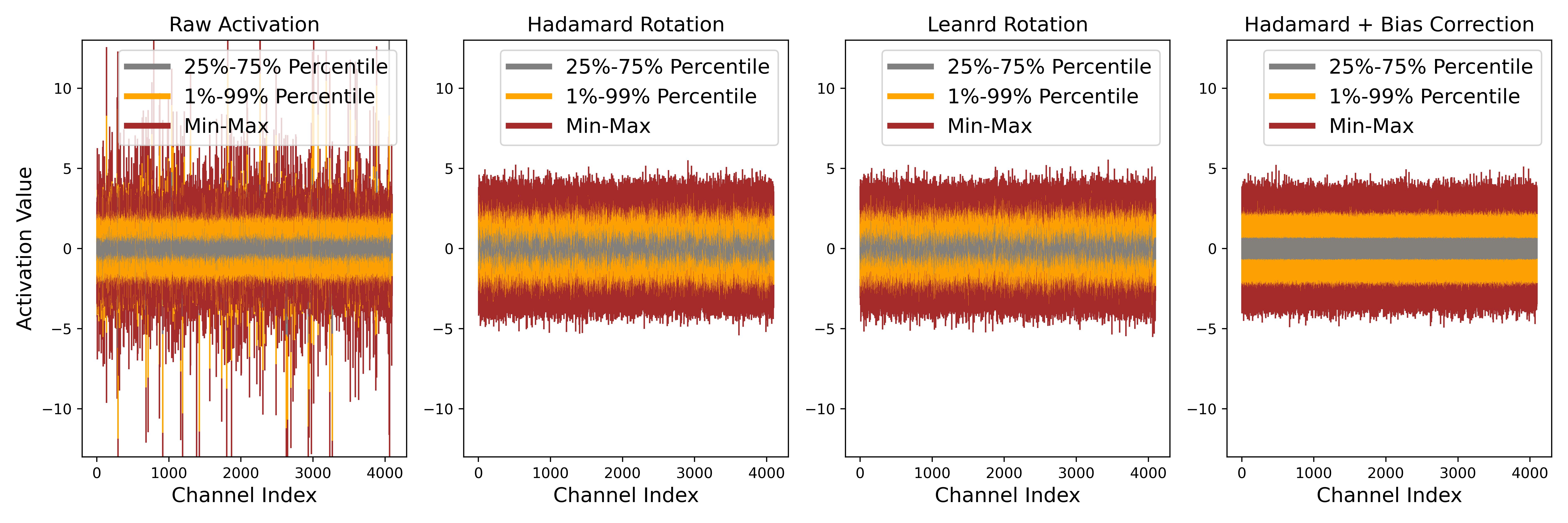}
	\caption{\small Visualizations comparing the activation distributions from the 11th MLP block in Llama3-8B.
}
        \end{center}
\end{figure} 

\begin{figure} [h]
	\begin{center}
	\includegraphics[width=1\textwidth]{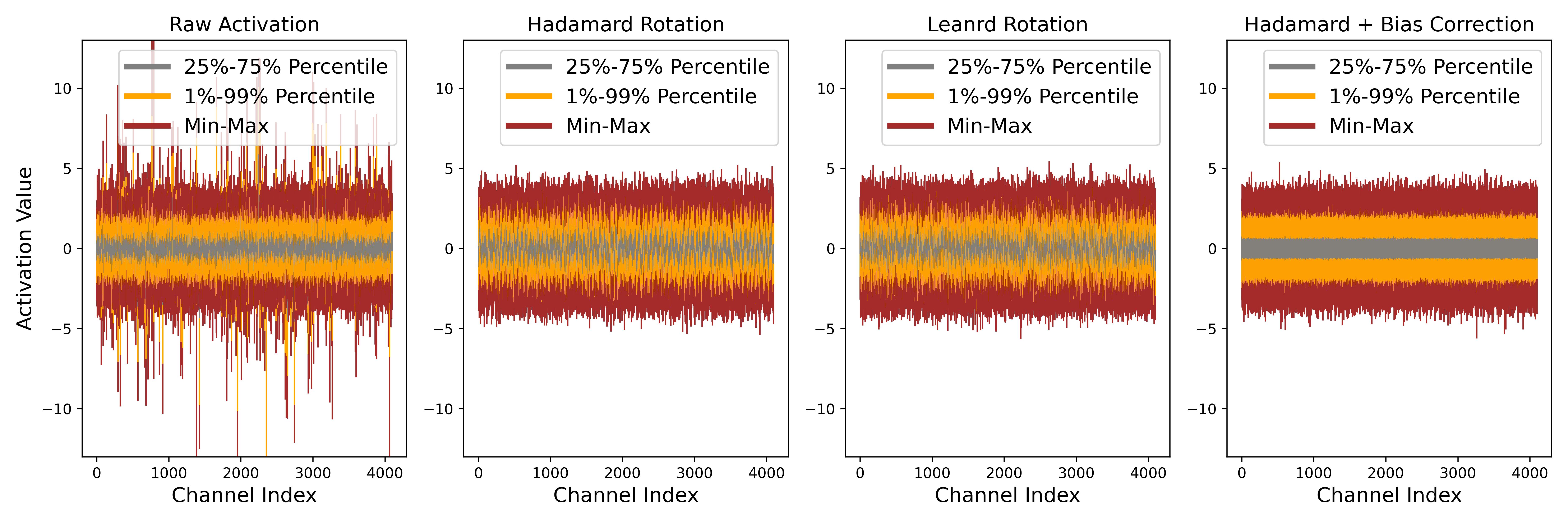}
	\caption{\small Visualizations comparing the activation distributions from the 31st MHSA block in Llama3-8B.
}
        \end{center}
\end{figure} 

\begin{figure} [h]
	\begin{center}
	\includegraphics[width=1\textwidth]{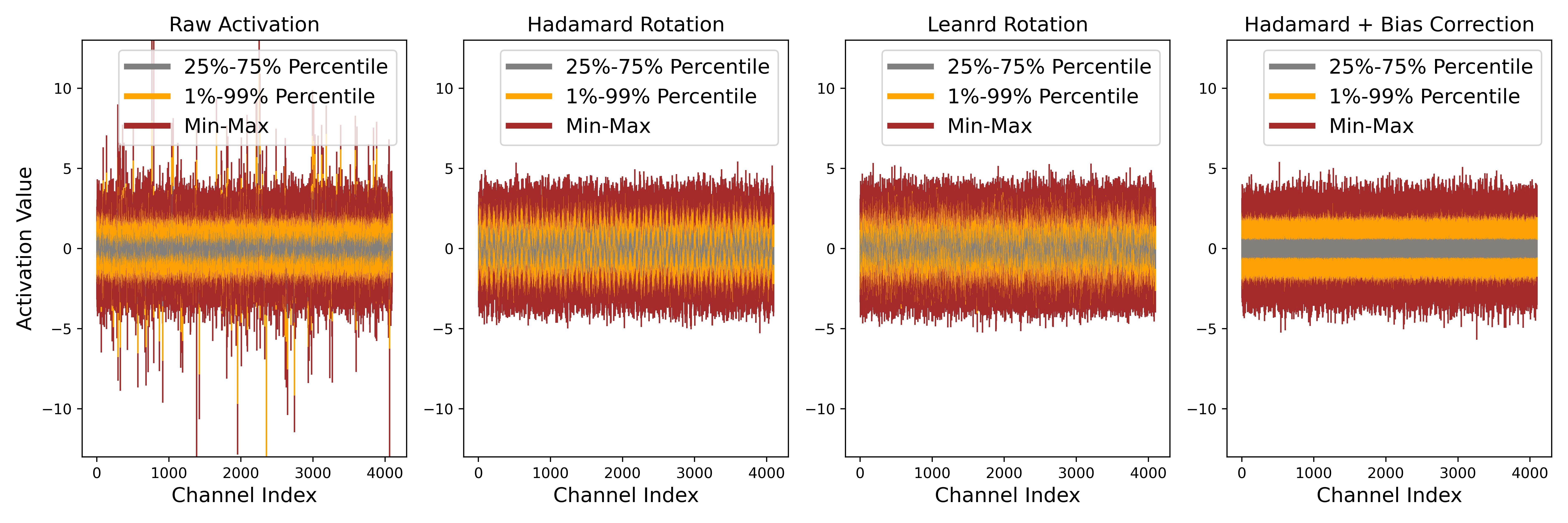}
	\caption{\small Visualizations comparing the activation distributions from the 31st MLP block in Llama3-8B.
}
        \end{center}
\end{figure} 

\clearpage
\section{Composition of rounding errors}
\label{appendix:rounding error}
This section presents visualizations of the composition of rounding errors, calculated according to \Cref{eq::5,eq::6}, across various models.

\begin{minipage}{0.45\linewidth}
  \centering
  \includegraphics[width=0.9\linewidth]{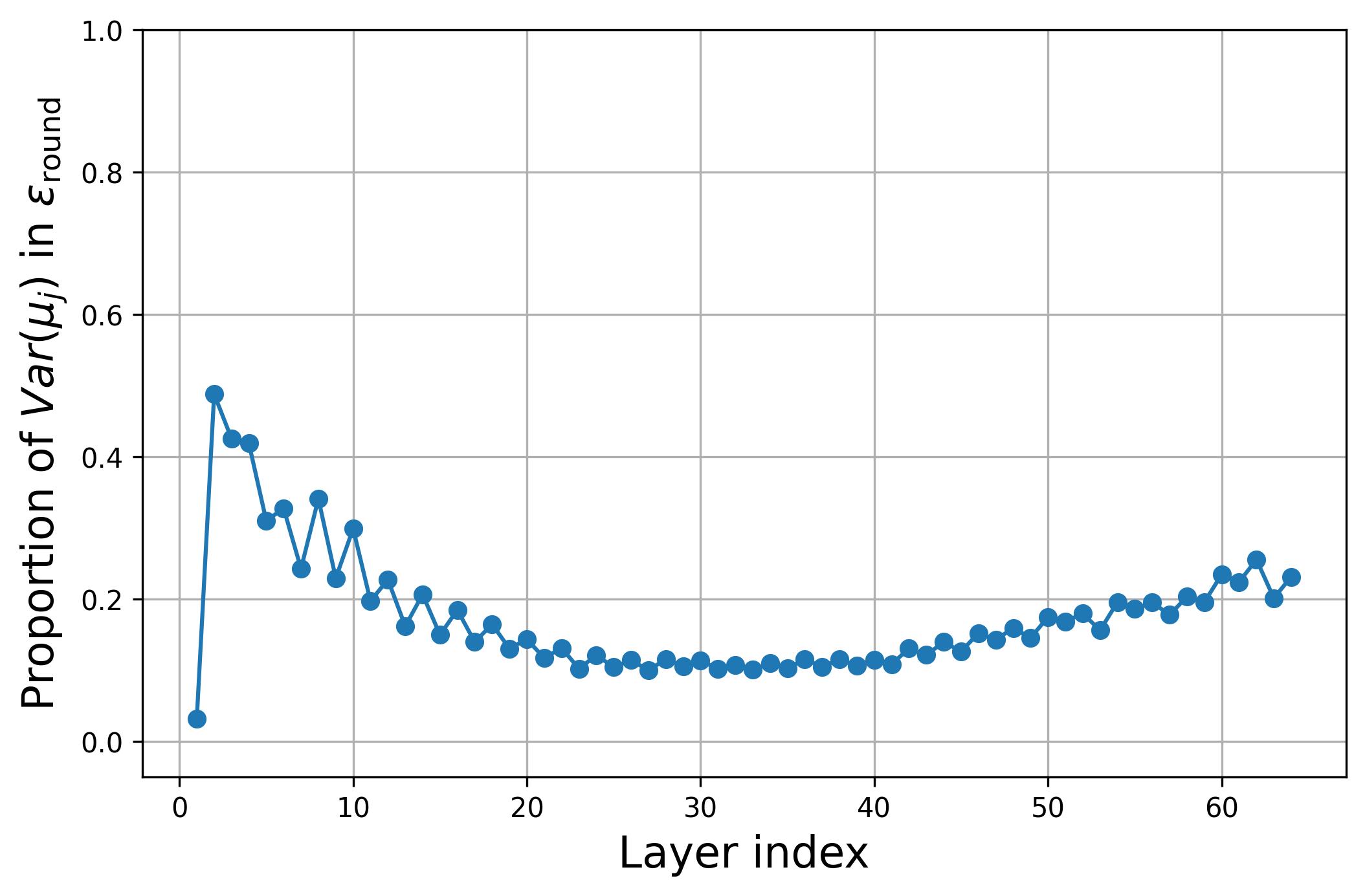}
  \captionof{figure}{Percentage of $Var(\mu_j)$ contribution to rounding error in Llama2-7B.}
\end{minipage}
\hfill
\begin{minipage}{0.45\linewidth}
  \centering
  \includegraphics[width=0.9\linewidth]{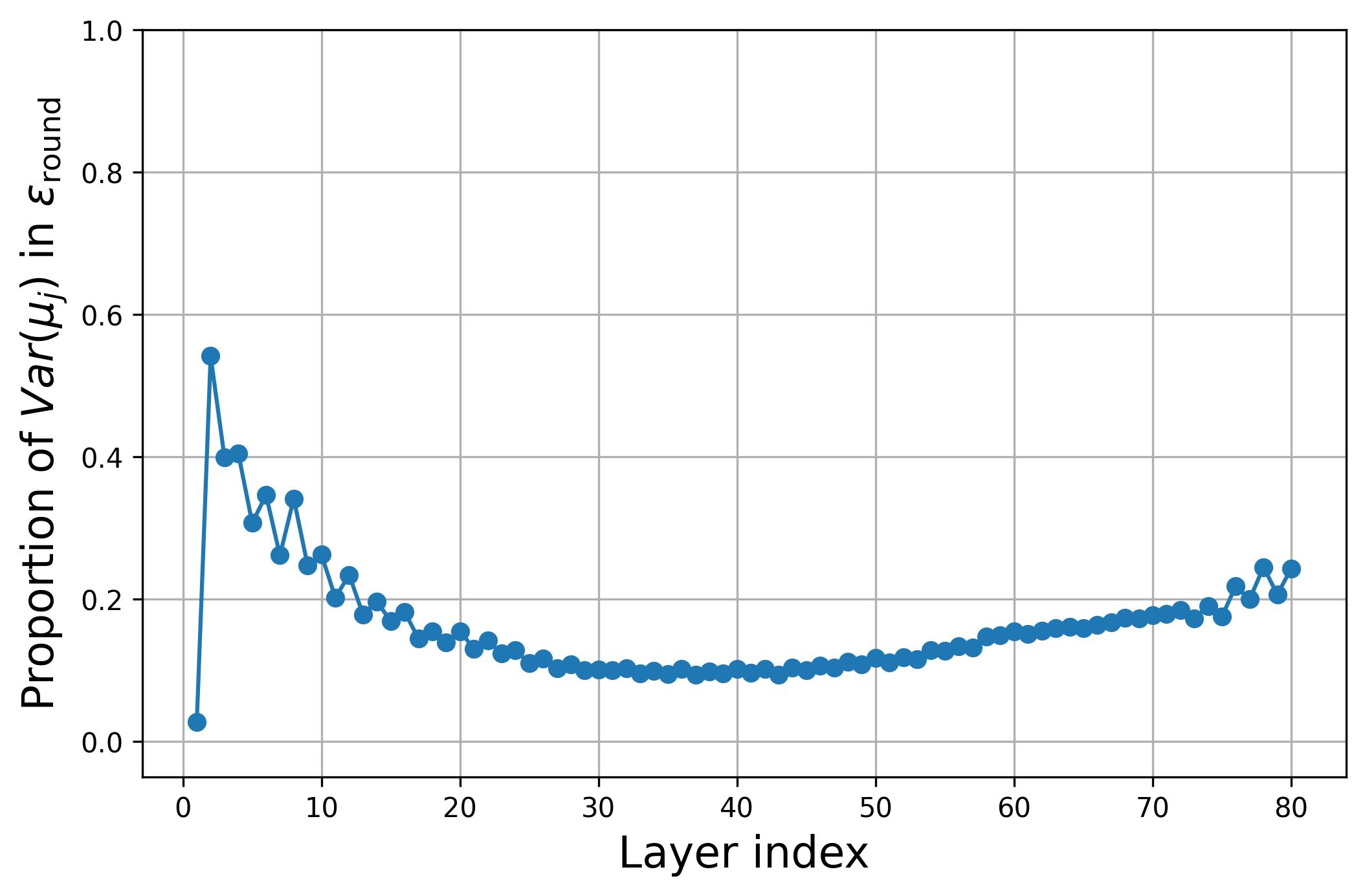}
  \captionof{figure}{Percentage of $Var(\mu_j)$ contribution to rounding error in Llama2-13B.}
\end{minipage}

\begin{minipage}{0.45\linewidth}
  \centering
  \includegraphics[width=0.9\linewidth]{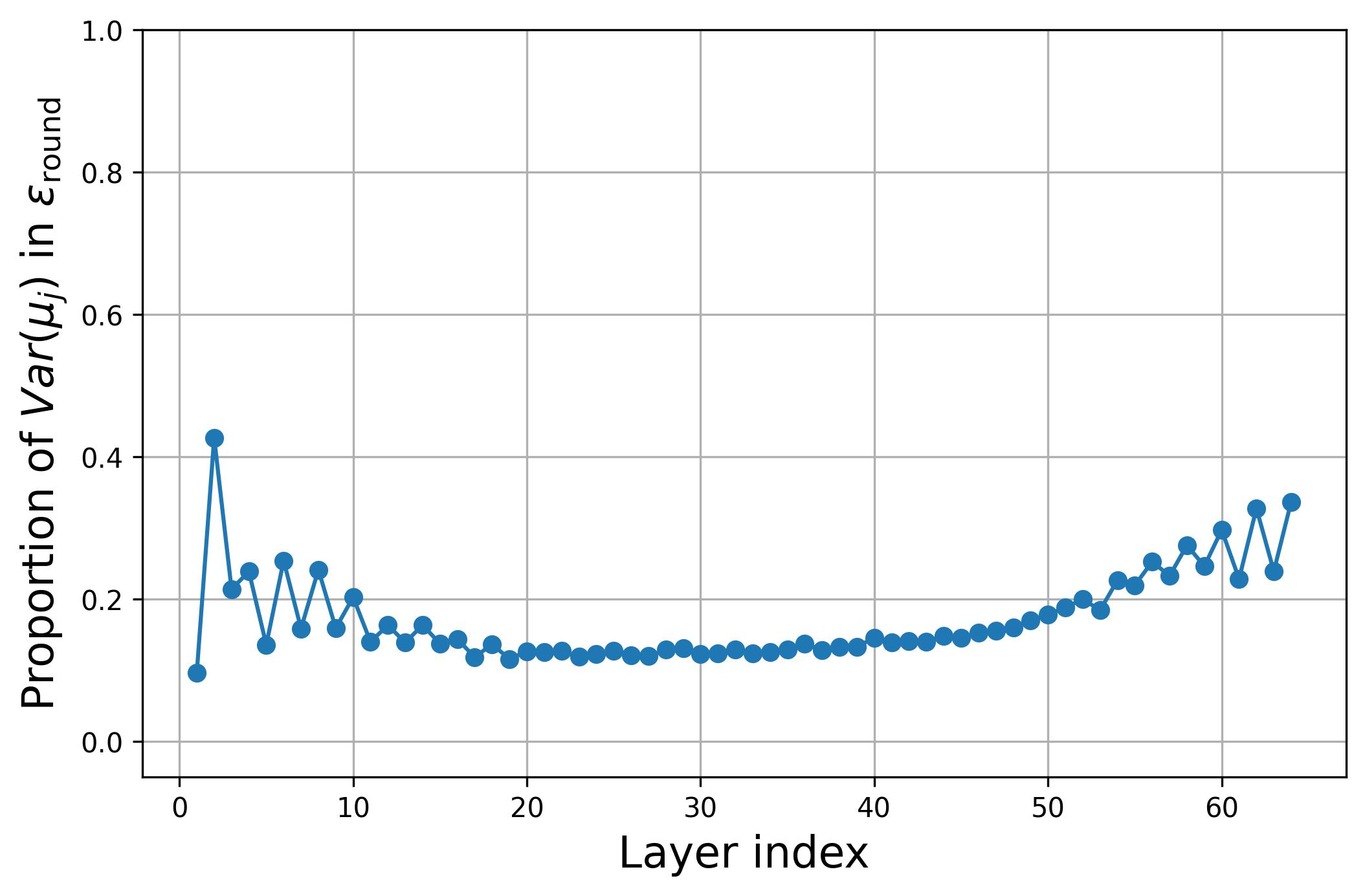}
  \captionof{figure}{Percentage of $Var(\mu_j)$ contribution to rounding error in Llama3-8B.}
\end{minipage}
\hfill
\begin{minipage}{0.45\linewidth}
  \centering
  \includegraphics[width=0.9\linewidth]{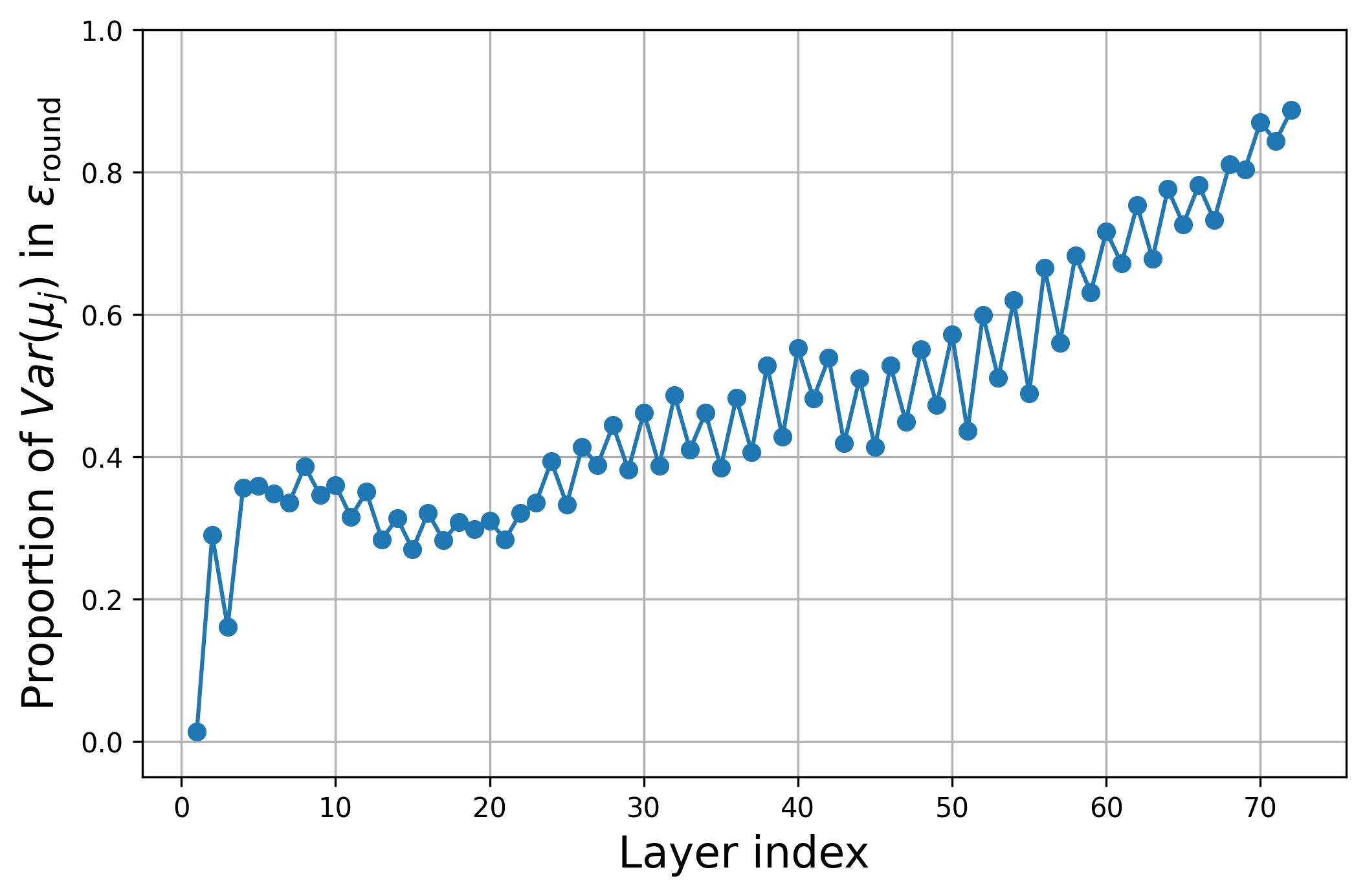}
  \captionof{figure}{Percentage of $Var(\mu_j)$ contribution to rounding error in Qwen2.5-3B.}
\end{minipage}

\begin{minipage}{0.45\linewidth}
  \centering
  \includegraphics[width=0.9\linewidth]{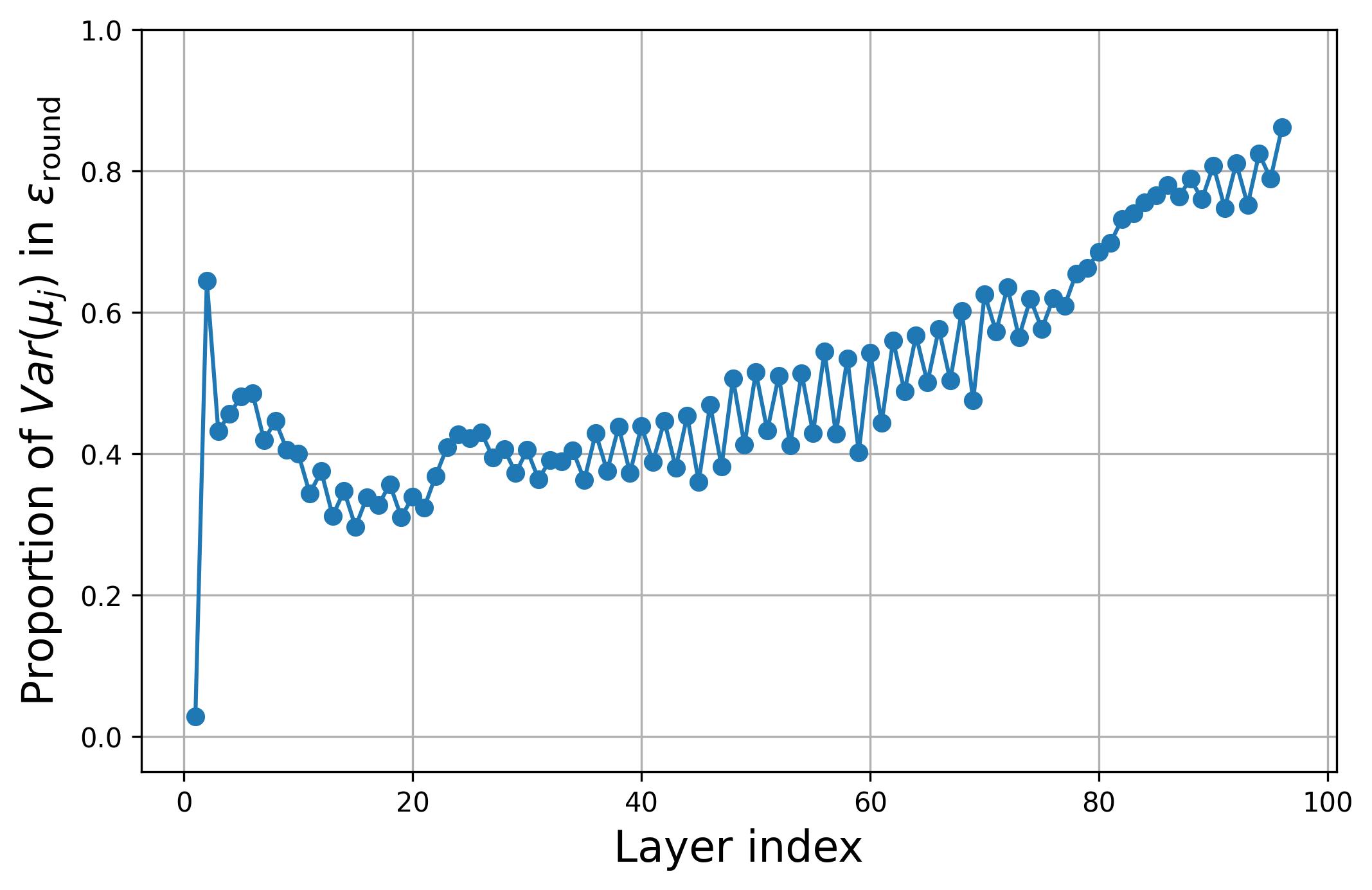}
  \captionof{figure}{Percentage of $Var(\mu_j)$ contribution to rounding error in Qwen2.5-14B.}
\end{minipage}
\hfill
\begin{minipage}{0.45\linewidth}
  \centering
  \includegraphics[width=0.9\linewidth]{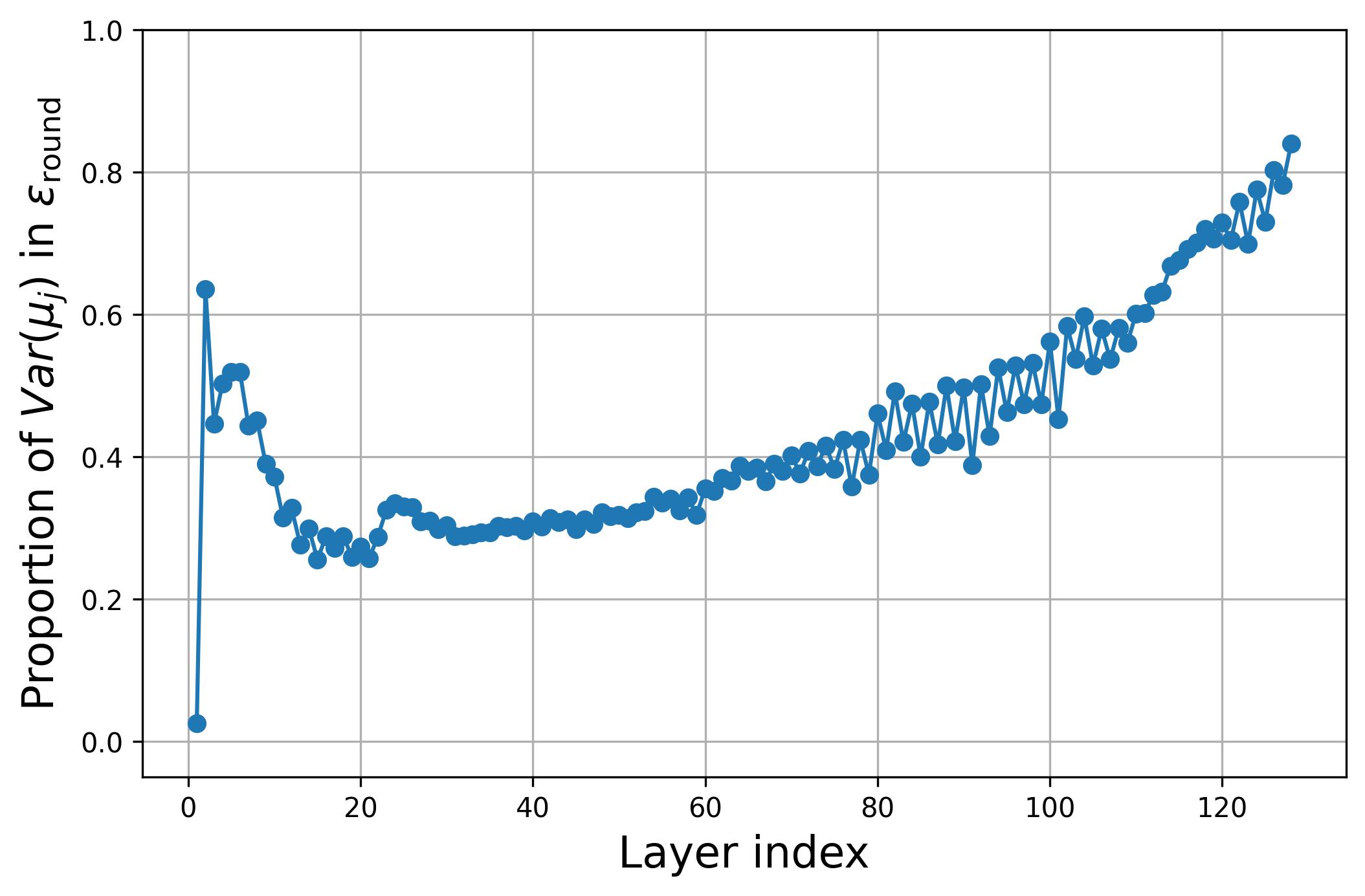}
  \captionof{figure}{Percentage of $Var(\mu_j)$ contribution to rounding error in Qwen2.5-32B.}
\end{minipage}

\end{document}

%% file: tables/appendix_all_results.tex
\begin{table*}[htbp]
\renewcommand\arraystretch{0.8}
\centering
\caption{\small Complete comparison of the perplexity score on WikiText2 and accuracy on Zero-shot Common Sense Reasoning tasks for \textbf{Llama-2 models}.}
\vspace{-3em}
\label{tab:llama2_comparison}
\setlength{\tabcolsep}{1mm}
{\resizebox{\textwidth}{!}{
\begin{tabular}{c|c|l|cccccccccc|c}
& & & & & & & & & & & &\\
& & & & & & & & & & & &\\
& & & & & & & & & & & &\\
& & & & & & & & & & & &\\
\noalign{\vspace{0.1em}}\hline\noalign{\vspace{0.1em}}
\noalign{\vspace{0.1em}}\hline\noalign{\vspace{0.2em}}
\multirow{2}{*}{\textbf{Model}} & \textbf{\#Bits} & \multirow{2}{*}{\textbf{Method}} & \textbf{ARC-c} & \textbf{ARC-e} & \textbf{BoolQ} & \textbf{HellaS.} & \textbf{Lam.} & \textbf{OBQA} & \textbf{PIQA} & \textbf{SIQA} & \textbf{WinoG.} & \textbf{Avg.} & \textbf{Wiki2} \\
& W-A-KV & & ($\uparrow$) & ($\uparrow$) & ($\uparrow$) & ($\uparrow$) & ($\uparrow$) & ($\uparrow$) & ($\uparrow$) & ($\uparrow$) & ($\uparrow$) & ($\uparrow$) & ($\downarrow$) \\
\noalign{\vspace{0.2em}}\hline\noalign{\vspace{0.2em}}
\multirow{5}{*}{2-7B} & 16-16-16 & Full Precision & 46.33 & 74.54 & 77.74 & 76.02 & 73.90 & 44.20 & 79.05 & 46.16 & 69.06 & 65.22 & 5.47 \\
\noalign{\vspace{0.2em}}\cdashline{2-14}\noalign{\vspace{0.2em}}
 & \multirow{4}{*}{4-4-4} & Quarot & 42.15 & 70.37 & 73.00 & 73.09 & 70.68 & 39.40 & 77.26 & 43.30 & 65.04 & 61.59 & 6.12 \\
 & & SpinQuant & 40.78 & 70.45 & 73.79 & 72.40 & 71.30 & 38.40 & 75.63 & 43.45 & 66.22 & 61.38 & 5.99 \\
 & & OSTQuant & 41.64 & 68.94 & 74.43 & 73.17 & 71.61 & \textbf{42.20} & 77.09 & \textbf{43.71} & 65.90 & 62.08 & 5.92 \\
 & & \cellcolor[rgb]{ .906, .902, .902}\textbf{BASE-Q} & \cellcolor[rgb]{ .906, .902, .902}\textbf{42.24} & \cellcolor[rgb]{ .906, .902, .902}\textbf{71.42} & \cellcolor[rgb]{ .906, .902, .902}\textbf{74.74} & \cellcolor[rgb]{ .906, .902, .902}\textbf{73.69} & \cellcolor[rgb]{ .906, .902, .902}\textbf{71.32} & \cellcolor[rgb]{ .906, .902, .902}41.80 & \cellcolor[rgb]{ .906, .902, .902}\textbf{77.53} & \cellcolor[rgb]{ .906, .902, .902}43.35 & \cellcolor[rgb]{ .906, .902, .902}\textbf{66.38} & \cellcolor[rgb]{ .906, .902, .902}\textbf{62.50} & \cellcolor[rgb]{ .906, .902, .902}\textbf{5.85} \\
\noalign{\vspace{0.2em}}\hline\noalign{\vspace{0.2em}}
\multirow{5}{*}{2-13B} & 16-16-16 & Full Precision & 49.15 & 77.44 & 80.61 & 79.38 & 76.73 & 45.20 & 80.52 & 47.39 & 72.14 & 67.62 & 4.88 \\
\noalign{\vspace{0.2em}}\cdashline{2-14}\noalign{\vspace{0.2em}}
 & \multirow{4}{*}{4-4-4} & Quarot & 46.76 & 75.08 & 76.97 & 75.84 & 74.36 & 42.80 & 78.84 & 45.19 & 69.46 & 65.03 & 5.39 \\
 & & SpinQuant & \textbf{48.46} & 74.33 & 77.28 & 76.27 & 74.83 & \textbf{44.60} & 78.67 & \textbf{46.01} & 70.24 & 65.63 & 5.30 \\
 & & OSTQuant & 47.10 & 75.20 & 77.46 & \textbf{77.71} & \textbf{75.14} & \textbf{44.60} & 78.67 & 45.75 & 68.03 & 65.41 & 5.24 \\
 & & \cellcolor[rgb]{ .906, .902, .902}\textbf{BASE-Q} & \cellcolor[rgb]{ .906, .902, .902}47.01 & \cellcolor[rgb]{ .906, .902, .902}\textbf{75.67} & \cellcolor[rgb]{ .906, .902, .902}\textbf{78.90} & \cellcolor[rgb]{ .906, .902, .902}77.42 & \cellcolor[rgb]{ .906, .902, .902}74.87 & \cellcolor[rgb]{ .906, .902, .902}\textbf{44.60} & \cellcolor[rgb]{ .906, .902, .902}\textbf{79.27} & \cellcolor[rgb]{ .906, .902, .902}45.34 & \cellcolor[rgb]{ .906, .902, .902}\textbf{70.48} & \cellcolor[rgb]{ .906, .902, .902}\textbf{65.95} & \cellcolor[rgb]{ .906, .902, .902}\textbf{5.19} \\
\noalign{\vspace{0.2em}}\hline\noalign{\vspace{0.2em}}
\multirow{5}{*}{2-70B} & 16-16-16 & Full Precision & 57.25 & 81.06 & 83.76 & 83.82 & 79.62 & 48.80 & 82.70 & 49.18 & 77.98 & 71.57 & 3.32 \\
\noalign{\vspace{0.2em}}\cdashline{2-14}\noalign{\vspace{0.2em}}
 & \multirow{4}{*}{4-4-4} & Quarot & 55.97 & \textbf{80.18} & 81.87 & 82.25 & 78.73 & 48.00 & 81.39 & 47.49 & 75.69 & 70.28 & 3.76 \\
 & & SpinQuant & 54.78 & 79.76 & 81.90 & 82.78 & 79.20 & 47.40 & 81.77 & 48.46 & \textbf{76.80} & 70.22 & 3.71 \\
 & & OSTQuant & OOM & OOM & OOM & OOM & OOM & OOM & OOM & OOM & OOM & OOM & OOM \\
 & & \cellcolor[rgb]{ .906, .902, .902}\textbf{BASE-Q} & \cellcolor[rgb]{ .906, .902, .902}\textbf{56.40} & \cellcolor[rgb]{ .906, .902, .902}80.09 & \cellcolor[rgb]{ .906, .902, .902}\textbf{82.17} & \cellcolor[rgb]{ .906, .902, .902}\textbf{82.79} & \cellcolor[rgb]{ .906, .902, .902}\textbf{79.37} & \cellcolor[rgb]{ .906, .902, .902}\textbf{48.20} & \cellcolor[rgb]{ .906, .902, .902}\textbf{82.32} & \cellcolor[rgb]{ .906, .902, .902}\textbf{48.62} & \cellcolor[rgb]{ .906, .902, .902}76.72 & \cellcolor[rgb]{ .906, .902, .902}\textbf{70.74} & \cellcolor[rgb]{ .906, .902, .902}\textbf{3.59} \\
\noalign{\vspace{0.2em}}\hline\noalign{\vspace{0.2em}}
\hline\noalign{\vspace{0.2em}}
\end{tabular}}}
\end{table*}

\begin{table*}[htbp]
\renewcommand\arraystretch{0.8}
\centering
\caption{\small Complete comparison of the perplexity score on WikiText2 and accuracy on Zero-shot Common Sense Reasoning tasks for \textbf{Llama-3 models}.}
\vspace{-4em}
\label{tab:llama3_comparison}
\setlength{\tabcolsep}{1mm}
{\resizebox{\textwidth}{!}{
\begin{tabular}{c|c|l|cccccccccc|c}
& & & & & & & & & & & &\\
& & & & & & & & & & & &\\
& & & & & & & & & & & &\\
& & & & & & & & & & & &\\
& & & & & & & & & & & &\\
& & & & & & & & & & & &\\
\noalign{\vspace{0.1em}}\hline\noalign{\vspace{0.1em}}
\noalign{\vspace{0.1em}}\hline\noalign{\vspace{0.2em}}
\multirow{2}{*}{\textbf{Model}} & \textbf{\#Bits} & \multirow{2}{*}{\textbf{Method}} & \textbf{ARC-c} & \textbf{ARC-e} & \textbf{BoolQ} & \textbf{HellaS.} & \textbf{Lam.} & \textbf{OBQA} & \textbf{PIQA} & \textbf{SIQA} & \textbf{WinoG.} & \textbf{Avg.} & \textbf{Wiki2} \\
& W-A-KV & & ($\uparrow$) & ($\uparrow$) & ($\uparrow$) & ($\uparrow$) & ($\uparrow$) & ($\uparrow$) & ($\uparrow$) & ($\uparrow$) & ($\uparrow$) & ($\uparrow$) & ($\downarrow$) \\
\noalign{\vspace{0.2em}}\hline\noalign{\vspace{0.2em}}
\multirow{5}{*}{3-8B} & 16-16-16 & Full Precision & 53.33 & 77.74 & 81.35 & 79.15 & 76.01 & 45.00 & 80.79 & 47.13 & 72.53 & 68.11 & 6.14 \\
\noalign{\vspace{0.2em}}\cdashline{2-14}\noalign{\vspace{0.2em}}
 & \multirow{4}{*}{4-4-4} & Quarot & 46.08 & 70.50 & 74.50 & 74.47 & 70.58 & 40.60 & 76.50 & 44.93 & 67.88 & 62.89 & 7.82 \\
 & & SpinQuant & 47.35 & 73.36 & 75.75 & 74.74 & 70.70 & 41.20 & 77.04 & 44.93 & 69.14 & 63.80 & 7.49 \\
 & & OSTQuant & 48.21 & 72.69 & \textbf{79.02} & 75.69 & 70.52 & \textbf{44.00} & 77.86 & 44.98 & 69.53 & 64.72 & 7.36 \\
 & & \cellcolor[rgb]{ .906, .902, .902}\textbf{BASE-Q} & \cellcolor[rgb]{ .906, .902, .902}\textbf{50.51} & \cellcolor[rgb]{ .906, .902, .902}\textbf{76.05} & \cellcolor[rgb]{ .906, .902, .902}78.26 & \cellcolor[rgb]{ .906, .902, .902}\textbf{76.48} & \cellcolor[rgb]{ .906, .902, .902}\textbf{71.22} & \cellcolor[rgb]{ .906, .902, .902}43.60 & \cellcolor[rgb]{ .906, .902, .902}\textbf{78.89} & \cellcolor[rgb]{ .906, .902, .902}\textbf{45.75} & \cellcolor[rgb]{ .906, .902, .902}\textbf{69.61} & \cellcolor[rgb]{ .906, .902, .902}\textbf{65.60} & \cellcolor[rgb]{ .906, .902, .902}\textbf{7.12} \\
\noalign{\vspace{0.2em}}\hline\noalign{\vspace{0.2em}}
\multirow{5}{*}{3-70B} & 16-16-16 & Full Precision & 64.33 & 85.90 & 85.23 & 84.89 & 79.82 & 48.60 & 84.55 & 50.72 & 80.35 & 73.82 & 2.86 \\
\noalign{\vspace{0.2em}}\cdashline{2-14}\noalign{\vspace{0.2em}}
 & \multirow{4}{*}{4-4-4} & Quarot & 53.16 & 78.11 & 82.91 & 80.80 & 75.49 & 44.40 & 79.65 & 47.49 & 76.87 & 68.76 & 5.62 \\
 & & SpinQuant & 57.25 & 80.22 & 83.06 & 81.05 & 75.04 & 46.20 & 81.88 & 47.44 & 77.27 & 69.93 & 5.11 \\
 & & OSTQuant & OOM & OOM & OOM & OOM & OOM & OOM & OOM & OOM & OOM & OOM & OOM \\
 & & \cellcolor[rgb]{ .906, .902, .902}\textbf{BASE-Q} & \cellcolor[rgb]{ .906, .902, .902}\textbf{59.98} & \cellcolor[rgb]{ .906, .902, .902}\textbf{82.70} & \cellcolor[rgb]{ .906, .902, .902}\textbf{84.65} & \cellcolor[rgb]{ .906, .902, .902}\textbf{83.92} & \cellcolor[rgb]{ .906, .902, .902}\textbf{78.42} & \cellcolor[rgb]{ .906, .902, .902}\textbf{47.20} & \cellcolor[rgb]{ .906, .902, .902}\textbf{83.08} & \cellcolor[rgb]{ .906, .902, .902}\textbf{48.41} & \cellcolor[rgb]{ .906, .902, .902}\textbf{78.14} & \cellcolor[rgb]{ .906, .902, .902}\textbf{71.83} & \cellcolor[rgb]{ .906, .902, .902}\textbf{4.06} \\
\noalign{\vspace{0.2em}}\hline\noalign{\vspace{0.2em}}
\multirow{5}{*}{3.1-8B} & 16-16-16 & Full Precision & 53.50 & 81.19 & 82.08 & 78.90 & 75.82 & 44.80 & 81.23 & 47.19 & 73.56 & 68.70 & 6.23 \\
\noalign{\vspace{0.2em}}\cdashline{2-14}\noalign{\vspace{0.2em}}
 & \multirow{4}{*}{4-4-4} & Quarot & 46.08 & 72.31 & 77.34 & 74.46 & 71.36 & 42.40 & 76.82 & 44.37 & 68.51 & 63.74 & 7.82 \\
 & & SpinQuant & 47.87 & 76.05 & 76.76 & 74.63 & 70.54 & \textbf{43.20} & \textbf{79.27} & 45.70 & 67.17 & 64.58 & 7.51 \\
 & & OSTQuant & 47.44 & 75.34 & \textbf{78.84} & 75.48 & \textbf{71.38} & 41.80 & 78.24 & 47.13 & 68.51 & 64.91 & 7.40 \\
 & & \cellcolor[rgb]{ .906, .902, .902}\textbf{BASE-Q} & \cellcolor[rgb]{ .906, .902, .902}\textbf{49.15} & \cellcolor[rgb]{ .906, .902, .902}\textbf{77.19} & \cellcolor[rgb]{ .906, .902, .902}78.56 & \cellcolor[rgb]{ .906, .902, .902}\textbf{76.41} & \cellcolor[rgb]{ .906, .902, .902}70.72 & \cellcolor[rgb]{ .906, .902, .902}42.40 & \cellcolor[rgb]{ .906, .902, .902}79.16 & \cellcolor[rgb]{ .906, .902, .902}\textbf{45.85} & \cellcolor[rgb]{ .906, .902, .902}\textbf{69.22} & \cellcolor[rgb]{ .906, .902, .902}\textbf{65.36} & \cellcolor[rgb]{ .906, .902, .902}\textbf{7.17} \\
\noalign{\vspace{0.2em}}\hline\noalign{\vspace{0.2em}}
\multirow{5}{*}{3.1-70B} & 16-16-16 & Full Precision & 64.93 & 86.66 & 85.41 & 85.02 & 79.16 & 48.00 & 84.28 & 50.56 & 79.79 & 73.76 & 2.81 \\
\noalign{\vspace{0.2em}}\cdashline{2-14}\noalign{\vspace{0.2em}}
 & \multirow{4}{*}{4-4-4} & Quarot & 57.51 & 80.35 & 83.27 & 81.45 & 75.65 & 44.80 & 81.66 & 45.60 & 75.77 & 69.56 & 5.31 \\
 & & SpinQuant & 59.13 & 82.28 & \textbf{84.37} & 82.24 & 76.32 & 46.00 & 82.21 & 47.24 & 76.40 & 70.69 & 4.74 \\
 & & OSTQuant & OOM & OOM & OOM & OOM & OOM & OOM & OOM & OOM & OOM & OOM & OOM \\
 & & \cellcolor[rgb]{ .906, .902, .902}\textbf{BASE-Q} & \cellcolor[rgb]{ .906, .902, .902}\textbf{60.49} & \cellcolor[rgb]{ .906, .902, .902}\textbf{84.13} & \cellcolor[rgb]{ .906, .902, .902}83.12 & \cellcolor[rgb]{ .906, .902, .902}\textbf{83.10} & \cellcolor[rgb]{ .906, .902, .902}\textbf{77.06} & \cellcolor[rgb]{ .906, .902, .902}\textbf{47.40} & \cellcolor[rgb]{ .906, .902, .902}\textbf{83.24} & \cellcolor[rgb]{ .906, .902, .902}\textbf{47.85} & \cellcolor[rgb]{ .906, .902, .902}\textbf{77.43} & \cellcolor[rgb]{ .906, .902, .902}\textbf{71.54} & \cellcolor[rgb]{ .906, .902, .902}\textbf{4.17} \\
\noalign{\vspace{0.2em}}\hline\noalign{\vspace{0.2em}}
\multirow{5}{*}{3.2-1B} & 16-16-16 & Full Precision & 36.35 & 60.48 & 64.07 & 63.65 & 62.93 & 37.20 & 74.54 & 42.99 & 60.77 & 55.89 & 9.75 \\
\noalign{\vspace{0.2em}}\cdashline{2-14}\noalign{\vspace{0.2em}}
 & \multirow{4}{*}{4-4-4} & Quarot & 31.23 & 51.01 & 59.42 & 54.74 & 43.90 & 34.40 & 66.92 & 40.48 & 55.88 & 48.66 & 14.44 \\
 & & SpinQuant & 32.68 & 51.68 & 58.47 & \textbf{56.68} & 47.22 & \textbf{34.80} & 67.79 & 40.23 & 55.17 & 49.41 & 13.46 \\
 & & OSTQuant & \textbf{33.45} & \textbf{55.26} & \textbf{61.80} & 56.02 & \textbf{48.90} & 33.80 & 70.02 & \textbf{41.61} & 56.83 & \textbf{50.85} & 12.84 \\
 & & \cellcolor[rgb]{ .906, .902, .902}\textbf{BASE-Q} & \cellcolor[rgb]{ .906, .902, .902}31.23 & \cellcolor[rgb]{ .906, .902, .902}53.91 & \cellcolor[rgb]{ .906, .902, .902}59.79 & \cellcolor[rgb]{ .906, .902, .902}56.61 & \cellcolor[rgb]{ .906, .902, .902}46.73 & \cellcolor[rgb]{ .906, .902, .902}31.60 & \cellcolor[rgb]{ .906, .902, .902}\textbf{70.57} & \cellcolor[rgb]{ .906, .902, .902}40.74 & \cellcolor[rgb]{ .906, .902, .902}\textbf{56.91} & \cellcolor[rgb]{ .906, .902, .902}49.79 & \cellcolor[rgb]{ .906, .902, .902}\textbf{12.63} \\
\noalign{\vspace{0.2em}}\hline\noalign{\vspace{0.2em}}
\multirow{5}{*}{3.2-3B} & 16-16-16 & Full Precision & 45.90 & 71.63 & 73.33 & 73.60 & 70.48 & 43.00 & 77.58 & 46.98 & 69.85 & 63.59 & 7.81 \\
\noalign{\vspace{0.2em}}\cdashline{2-14}\noalign{\vspace{0.2em}}
 & \multirow{4}{*}{4-4-4} & Quarot & 38.40 & 59.13 & 64.56 & 66.49 & 60.00 & 37.20 & 72.25 & 42.68 & 62.04 & 55.86 & 10.07 \\
 & & SpinQuant & 37.12 & 60.90 & 68.72 & 68.83 & 61.67 & 39.60 & 73.88 & 44.58 & 64.40 & 57.74 & 9.32 \\
 & & OSTQuant & 41.04 & \textbf{68.06} & 68.84 & 68.75 & \textbf{63.50} & 40.40 & 74.21 & \textbf{44.78} & 64.01 & 59.29 & 9.16 \\
 & & \cellcolor[rgb]{ .906, .902, .902}\textbf{BASE-Q} & \cellcolor[rgb]{ .906, .902, .902}\textbf{41.64} & \cellcolor[rgb]{ .906, .902, .902}66.86 & \cellcolor[rgb]{ .906, .902, .902}\textbf{72.45} & \cellcolor[rgb]{ .906, .902, .902}\textbf{69.83} & \cellcolor[rgb]{ .906, .902, .902}62.57 & \cellcolor[rgb]{ .906, .902, .902}\textbf{40.80} & \cellcolor[rgb]{ .906, .902, .902}\textbf{75.84} & \cellcolor[rgb]{ .906, .902, .902}44.52 & \cellcolor[rgb]{ .906, .902, .902}\textbf{65.67} & \cellcolor[rgb]{ .906, .902, .902}\textbf{60.02} & \cellcolor[rgb]{ .906, .902, .902}\textbf{9.01} \\
\noalign{\vspace{0.2em}}\hline\noalign{\vspace{0.2em}}
\hline\noalign{\vspace{0.2em}}
\end{tabular}}}
\end{table*}

\begin{table*}[htbp]
\renewcommand\arraystretch{0.8}
\centering
\caption{\small Complete comparison of the perplexity score on WikiText2 and accuracy on Zero-shot Common Sense Reasoning tasks for \textbf{Qwen2.5 Models}.}
\vspace{-3em}
\label{tab:qwen25_comparison}
\setlength{\tabcolsep}{1mm}
{\resizebox{\textwidth}{!}{
\begin{tabular}{c|c|l|cccccccccc|c}
& & & & & & & & & & & &\\
& & & & & & & & & & & &\\
& & & & & & & & & & & &\\
& & & & & & & & & & & &\\
\noalign{\vspace{0.1em}}\hline\noalign{\vspace{0.1em}}
\noalign{\vspace{0.1em}}\hline\noalign{\vspace{0.2em}}
\multirow{2}{*}{\textbf{Model}} & \textbf{\#Bits} & \multirow{2}{*}{\textbf{Method}} & \textbf{ARC-c} & \textbf{ARC-e} & \textbf{BoolQ} & \textbf{HellaS.} & \textbf{Lam.} & \textbf{OBQA} & \textbf{PIQA} & \textbf{SIQA} & \textbf{WinoG.} & \textbf{Avg.} & \textbf{Wiki2} \\
& W-A-KV & & ($\uparrow$) & ($\uparrow$) & ($\uparrow$) & ($\uparrow$) & ($\uparrow$) & ($\uparrow$) & ($\uparrow$) & ($\uparrow$) & ($\uparrow$) & ($\uparrow$) & ($\downarrow$) \\
\noalign{\vspace{0.2em}}\hline\noalign{\vspace{0.2em}}
\multirow{5}{*}{2.5-3B} & 16-16-16 & Full Precision & 47.53 & 73.06 & 77.22 & 73.52 & 67.11 & 42.00 & 78.84 & 49.80 & 64.14 & 64.17 & 8.03 \\
\noalign{\vspace{0.2em}}\cdashline{2-14}\noalign{\vspace{0.2em}}
 & \multirow{4}{*}{4-4-4} & Quarot & 30.72 & 52.69 & 51.16 & 49.52 & 20.63 & 34.60 & 66.70 & 38.84 & 53.83 & 44.30 & 69.33 \\
 & & SpinQuant & 34.47 & 57.58 & 54.04 & 50.91 & 24.14 & 32.20 & 68.01 & 40.48 & 57.93 & 46.86 & 46.35 \\
 & & OSTQuant & 38.48 & 58.08 & 61.90 & 56.19 & 36.33 & 37.60 & 67.90 & 42.48 & 58.33 & 50.81 & 20.09 \\
 & & \cellcolor[rgb]{ .906, .902, .902}\textbf{BASE-Q} & \cellcolor[rgb]{ .906, .902, .902}\textbf{42.75} & \cellcolor[rgb]{ .906, .902, .902}\textbf{67.26} & \cellcolor[rgb]{ .906, .902, .902}\textbf{71.96} & \cellcolor[rgb]{ .906, .902, .902}\textbf{66.21} & \cellcolor[rgb]{ .906, .902, .902}\textbf{52.24} & \cellcolor[rgb]{ .906, .902, .902}\textbf{38.40} & \cellcolor[rgb]{ .906, .902, .902}\textbf{74.59} & \cellcolor[rgb]{ .906, .902, .902}\textbf{46.01} & \cellcolor[rgb]{ .906, .902, .902}\textbf{63.77} & \cellcolor[rgb]{ .906, .902, .902}\textbf{58.13} & \cellcolor[rgb]{ .906, .902, .902}\textbf{10.83} \\
\noalign{\vspace{0.2em}}\hline\noalign{\vspace{0.2em}}
\multirow{5}{*}{2.5-14B} & 16-16-16 & Full Precision & 58.87 & 79.17 & 85.26 & 82.91 & 74.62 & 45.20 & 82.05 & 55.17 & 75.30 & 70.95 & 5.29 \\
\noalign{\vspace{0.2em}}\cdashline{2-14}\noalign{\vspace{0.2em}}
 & \multirow{4}{*}{4-4-4} & Quarot & 55.80 & 79.46 & 80.24 & 78.40 & 70.06 & 41.60 & 78.45 & 50.20 & 70.88 & 67.23 & 6.77 \\
 & & SpinQuant & 53.75 & 79.97 & 78.87 & 79.18 & \textbf{70.97} & 44.00 & 79.22 & 49.23 & 70.40 & 67.29 & 6.55 \\
 & & OSTQuant & 54.78 & 78.91 & 80.73 & \textbf{79.81} & 69.18 & \textbf{44.40} & 79.71 & 50.15 & 72.61 & 67.81 & 6.37 \\
 & & \cellcolor[rgb]{ .906, .902, .902}\textbf{BASE-Q} & \cellcolor[rgb]{ .906, .902, .902}\textbf{56.14} & \cellcolor[rgb]{ .906, .902, .902}\textbf{82.41} & \cellcolor[rgb]{ .906, .902, .902}\textbf{82.23} & \cellcolor[rgb]{ .906, .902, .902}79.75 & \cellcolor[rgb]{ .906, .902, .902}70.89 & \cellcolor[rgb]{ .906, .902, .902}42.80 & \cellcolor[rgb]{ .906, .902, .902}\textbf{80.20} & \cellcolor[rgb]{ .906, .902, .902}\textbf{52.97} & \cellcolor[rgb]{ .906, .902, .902}\textbf{72.69} & \cellcolor[rgb]{ .906, .902, .902}\textbf{68.90} & \cellcolor[rgb]{ .906, .902, .902}\textbf{6.28} \\
\noalign{\vspace{0.2em}}\hline\noalign{\vspace{0.2em}}
\multirow{5}{*}{2.5-32B} & 16-16-16 & Full Precision & 55.63 & 80.93 & 87.19 & 84.06 & 76.96 & 44.00 & 82.32 & 56.29 & 75.22 & 71.11 & 5.02 \\
\noalign{\vspace{0.2em}}\cdashline{2-14}\noalign{\vspace{0.2em}}
 & \multirow{4}{*}{4-4-4} & Quarot & 52.05 & 74.66 & 85.41 & 81.70 & 73.14 & 42.40 & 79.98 & 52.81 & 71.11 & 68.14 & 6.04 \\
 & & SpinQuant & 52.82 & 76.05 & 85.11 & 80.99 & 72.71 & 44.40 & 80.03 & 52.15 & 72.30 & 68.51 & 5.88 \\
 & & OSTQuant & OOM & OOM & OOM & OOM & OOM & OOM & OOM & OOM & OOM & OOM & OOM \\
 & & \cellcolor[rgb]{ .906, .902, .902}\textbf{BASE-Q} & \cellcolor[rgb]{ .906, .902, .902}\textbf{54.61} & \cellcolor[rgb]{ .906, .902, .902}\textbf{78.16} & \cellcolor[rgb]{ .906, .902, .902}\textbf{87.06} & \cellcolor[rgb]{ .906, .902, .902}\textbf{82.49} & \cellcolor[rgb]{ .906, .902, .902}\textbf{75.61} & \cellcolor[rgb]{ .906, .902, .902}\textbf{44.60} & \cellcolor[rgb]{ .906, .902, .902}\textbf{81.12} & \cellcolor[rgb]{ .906, .902, .902}\textbf{53.63} & \cellcolor[rgb]{ .906, .902, .902}\textbf{74.35} & \cellcolor[rgb]{ .906, .902, .902}\textbf{70.18} & \cellcolor[rgb]{ .906, .902, .902}\textbf{5.65} \\
\noalign{\vspace{0.2em}}\hline\noalign{\vspace{0.2em}}
\hline\noalign{\vspace{0.2em}}
\end{tabular}}}
\end{table*}